\documentclass{article}

\usepackage[numbers]{natbib}
\usepackage{xcolor}
\definecolor{darkgreen}{RGB}{0, 150, 0}
\definecolor{urlgreen}{RGB}{0, 100, 0}
\definecolor{eccvblue}{RGB}{33, 79, 173}
\usepackage[colorlinks=true, citecolor=darkgreen, linkcolor=red, urlcolor=eccvblue]{hyperref}
\usepackage[preprint]{neurips_2026}

\usepackage[utf8]{inputenc} 
\usepackage[T1]{fontenc}    
\usepackage{url}            
\usepackage{booktabs}       
\usepackage{amsfonts}       
\usepackage{nicefrac}       
\usepackage{microtype}      
\usepackage{xcolor}         
\usepackage{colortbl} 
\usepackage{amsmath}
\usepackage{graphicx}
\usepackage{multirow}
\usepackage{caption}
\usepackage{placeins}
\usepackage{makecell}
\usepackage{subcaption}
\definecolor{bestrow}{RGB}{220, 240, 220} 
\begin{document}
\title{OTT-Vid: Optimal Transport Temporal Token Compression 
for Video Large Language Models}

%

\author{%
  Minseok Kang\thanks{Email: \texttt{louis0503@yonsei.ac.kr}} \\
  Yonsei University \\
  \And
  Minhyeok Lee \\
  Yonsei University \\
  \And
  Jungho Lee \\
  Yonsei University \\
  \And
  Minjung Kim \\
  LG Electronics \\
  \AND
  Donghyeong Kim \\
  Yonsei University \\
  \And
  Dayeon Lee \\
  Yonsei University \\
  \And
  Heeseung Choi \\
  KIST \\
  \And
  Ig-Jae Kim \\
  KIST \\
  \And
  Sangyoun Lee\thanks{Corresponding author.} \\
  Yonsei University \\
}

\maketitle

\begin{center}
\vspace{-0.8cm}
\texttt{\url{https://github.com/minseokii/OTT-Vid}}
\vspace{0.1cm}
\end{center}

\begin{abstract}
As Video Large Language Models (Video-LLMs) scale to longer
and more complex videos, their inference cost grows rapidly
due to the large volume of visual tokens accumulated across
frames. Training-free token compression has emerged as a
practical solution to this bottleneck. However, existing
temporal compression methods rely primarily on cross-frame
token similarity or segmentation heuristics, overlooking each
token's semantic role within its frame and failing to adapt
compression strength to the compressibility of each frame
pair. In this work, we propose OTT-Vid, a transport-derived
allocation framework for temporal token compression. Our
approach consists of two stages: spatial pruning identifies
representative content within each frame, and optimal
transport (OT) is then solved between neighboring frames to
estimate temporal compressibility. We formulate this OT with
non-uniform token mass, which protects semantically important
tokens from aggressive compression, and a locality-aware cost
that captures both feature and spatial disparities. The
resulting transport plan jointly balances token importance
and matching cost, while its total cost defines the transport
difficulty of each frame pair, which we use to allocate
compression budgets dynamically. Experiments on six benchmarks
spanning video question answering and temporal grounding show
that OTT-Vid preserves 95.8\% of VQA and 73.9\% of VTG
performance while retaining only 10\% of tokens, consistently
outperforming existing state-of-the-art training-free
compression methods.
\end{abstract}

\section{Introduction}
Recent advances in Video-LLMs~\cite{llava-video, llava-ov, qwen2vl, qwen2.5vl} have achieved remarkable progress in video understanding, enabling complex tasks such as video question answering, temporal grounding, and long video reasoning. As these capabilities extend to longer and more complex videos, a practical bottleneck has become increasingly apparent: processing many frames requires a large number of visual tokens, which quickly drives up sequence length and computational cost. In practice, the resulting visual token overhead often limits how many frames a Video-LLM can handle effectively.

To mitigate this bottleneck, training-free token compression 
has emerged as a practical direction that preserves 
compatibility with pretrained Video-LLMs without retraining. Such methods can be broadly categorized into two regimes according to where compression is applied in the inference pipeline: Inner-LLM methods and Outer-LLM methods. Inner-LLM methods~\cite{fastv, framefusion, sttm} operate within the transformer layers of the LLM to prune or merge tokens at intermediate depths. While they benefit from richer multimodal and query-aware signals for compression, these signals become available only during prefill, limiting the overall efficiency gains. Outer-LLM methods~\cite{visionzip, fastvid, flashvid, forestprune, floc}, by contrast, compress visual tokens before they enter the LLM, directly shortening the sequence length processed by the model. Outer-LLM compression is therefore more effective at reducing the cost of handling long visual sequences.
Within the outer-LLM regime, video token compression 
targets two sources of redundancy under a fixed token 
budget: spatial redundancy, where nearby patches within 
a frame encode overlapping content, and temporal 
redundancy, where similar content recurs across 
neighboring frames. Since current Video-LLMs encode each frame independently 
through an image encoder, spatial redundancy 
can be identified using per-frame signals such as 
attention-based saliency~\cite{visionzip, holitom}, 
diversity~\cite{divprune, prunesid}, 
coverage~\cite{scope, floc}, 
and intra-frame feature similarity~\cite{sttm, flashvid, fastvid}. 
Temporal redundancy, by contrast, lacks explicit cross-frame cues 
from the encoder, and therefore most existing 
methods~\cite{sttm, holitom, flashvid, fastvid, forestprune} 
rely primarily on token similarity across adjacent frames.

\begin{figure*}[t]
\centering
\includegraphics[width=\textwidth]{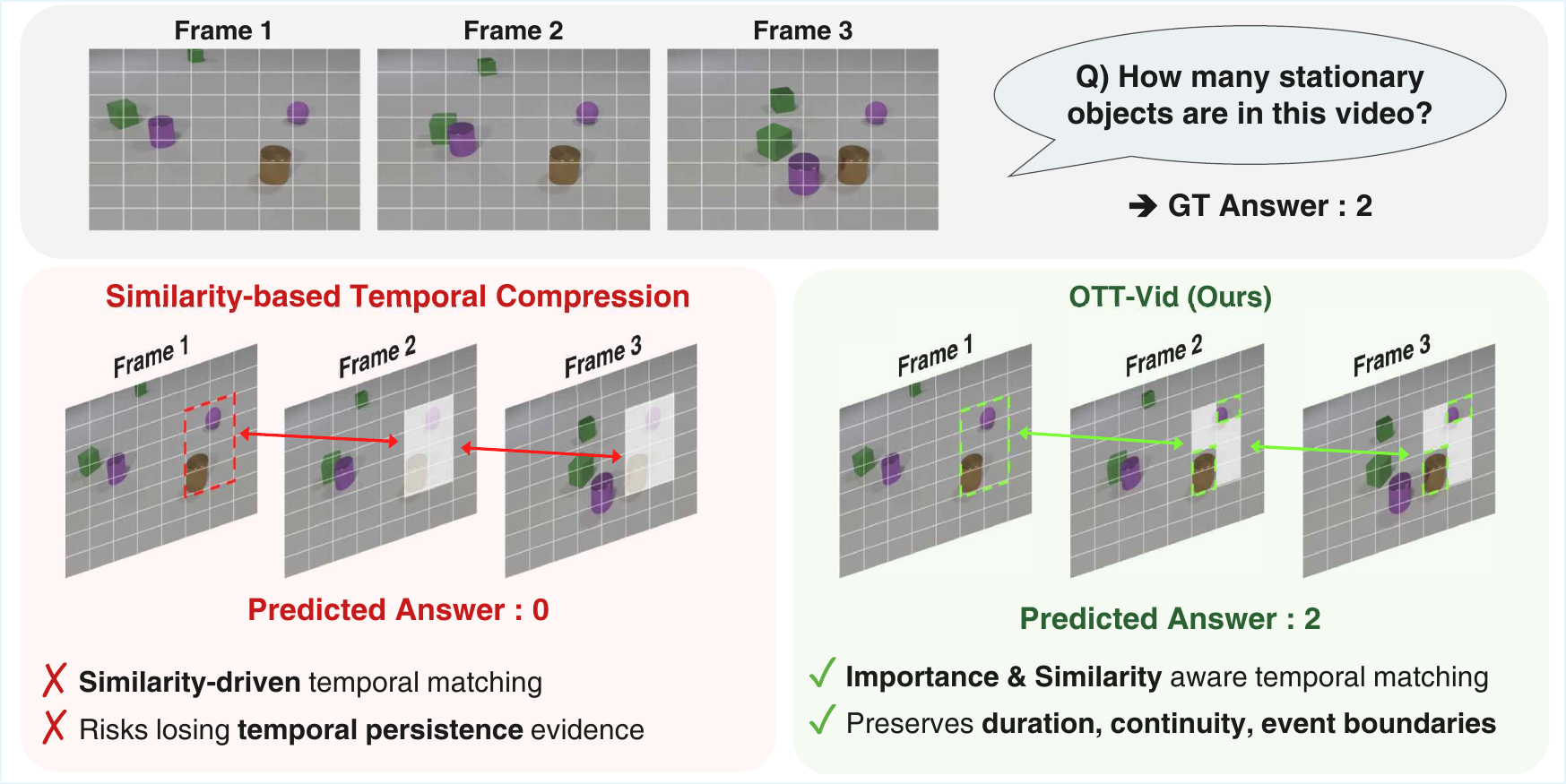}
\caption{Illustration of temporal compression strategies, where white-shaded regions denote compressed tokens. For simplicity, only compression of stationary object regions is depicted. Similarity-driven compression merges stationary objects away due to high 
cross-frame similarity, losing persistence evidence. 
OTT-Vid jointly considers token importance and 
similarity, preserving semantically representative tokens 
across frames.}
\vspace{-0.4cm}
\label{fig:teaser}
\end{figure*}

However, high token similarity across neighboring frames does not necessarily imply redundancy. Temporal repetition often reflects the persistence of semantically important content, which similarity alone cannot distinguish from uninformative background recurrence. Temporal compression driven purely by similarity therefore risks losing evidence of duration, continuity, and event boundaries. Figure~\ref{fig:teaser} illustrates this with stationary objects: although their temporal persistence constitutes essential evidence for the correct answer, their high cross-frame similarity can cause similarity-driven compression to discard them as redundant. In contrast, an importance-aware strategy compresses uninformative regions while preserving such persistence evidence, leading to the correct prediction. The same failure mode manifests at the task level in temporal grounding, where existing similarity-driven methods~\cite{fastvid, flashvid} exhibit substantially larger performance drops relative to standard video question answering, reflecting their inability to preserve the temporal evidence such tasks rely on. This motivates incorporating intra-frame token importance into temporal
compression, providing a semantic prior that helps separate meaningful
persistence from removable redundancy.

A further challenge is that temporal compression requires 
not only identifying which cross-frame correspondences 
are safely compressible, but also deciding how much 
compression each frame pair can tolerate under a limited 
token budget. Temporal redundancy is not distributed 
uniformly across a video: some neighboring frames are 
highly redundant and can tolerate aggressive compression, 
whereas others contain temporally informative evidence 
that should be preserved more carefully. Existing methods, 
however, do not treat this as an explicit allocation 
problem, instead relying on similarity thresholds~\cite{sttm,holitom} or hard 
temporal boundaries~\cite{flashvid,fastvid} that do not 
adapt to the compressibility of each frame pair. This can 
lead to under-compression of highly redundant regions and 
over-compression of regions whose temporal evidence 
should be preserved.

We propose \textbf{OTT-Vid}, a training-free video token 
compression framework that first applies spatial pruning to 
retain representative semantic content within each frame, 
and then performs adaptive temporal compression between 
neighboring frames. Our method decomposes temporal 
compression into two coupled questions: \textbf{(1)} how 
much information each token should preserve, and 
\textbf{(2)} how easily this information can be transferred 
across neighboring frames. Optimal transport (OT) is a principled 
framework for solving these two questions jointly, as its 
two primitives, \textbf{mass} and \textbf{cost}, are directly 
suited to encode token preservation priority and 
cross-frame transferability, respectively. 
In OT, the mass of each token specifies how much of it 
must be transferred. Viewing transport as compression, since larger mass is transferred more extensively, mass directly encodes 
compression priority. Compressing while preserving 
important content thus calls for a counter-intuitive 
design that assigns smaller mass to more important 
tokens. We therefore set mass inversely to frame-wise 
token importance, so that semantically representative 
tokens receive smaller mass and resist compression, while 
less informative tokens receive larger mass and are more 
readily transferred. The cost between two tokens 
quantifies the difficulty of mass transfer, and we define 
it from spatial distance and feature dissimilarity so that 
nearby and similar tokens incur low cost while distant or 
dissimilar pairs incur high cost. By solving a single 
optimization problem with these two quantities, OT yields 
a transport plan for each adjacent frame pair that 
jointly balances the two questions in a unified solution.
 The resulting transport plan characterizes which token correspondences carry high compression priority, together with a transport difficulty that signals how much redundancy the pair contains. We use this 
difficulty to distribute the compression budget non-uniformly 
across frame pairs, and execute compression within each 
pair according to its transport plan and allocated budget. 
In this way, OTT-Vid unifies token-level compression 
decisions with pair-level budget allocation in a single 
transport formulation.

We evaluate OTT-Vid on four video question answering 
benchmarks (MVBench~\cite{mvbench}, VideoMME~\cite{videomme}, 
LongVideoBench~\cite{longvideobench}, MLVU~\cite{mlvu}) 
and two video temporal grounding benchmarks (Charades-STA~\cite{charades}, 
ActivityNet-Captions~\cite{activitynet}, with refined 
annotations from TimeLens~\cite{timelens}). The latter directly tests whether compression preserves temporal evidence, which prior work rarely evaluates. On Qwen2.5-VL-7B~\cite{qwen2.5vl}, OTT-Vid consistently outperforms strong training-free baselines under matched token budgets, preserving 95.8\% of uncompressed VQA performance and 73.9\% of VTG performance at 10\% token retention. Experiments on LLaVA-OV~\cite{llava-ov} and LLaVA-Video~\cite{llava-video} confirm that these gains generalize across Video-LLM backbones.

\section{Background and Related Work}
\subsection{Video-LLM Inference Pipeline}

Recent Video-LLMs such as LLaVA-OneVision~\cite{llava-ov}, 
LLaVA-Video~\cite{llava-video}, and 
Qwen2.5-VL~\cite{qwen2.5vl} combine a pretrained image 
encoder, a lightweight projector, and a large language 
model. Each frame is encoded independently into $N_v$ 
visual tokens and projected into the language embedding 
space to form $H_v \in \mathbb{R}^{T \cdot N_v \times d}$. 
A text prompt is embedded as 
$H_q \in \mathbb{R}^{N_q \times d}$, and the multimodal 
input $H = \mathrm{concat}(H_v, H_q)$ has total length 
$n = T \cdot N_v + N_q$. The prefilling stage dominates 
inference cost due to quadratic self-attention over this 
sequence, and since $T \cdot N_v$ typically far exceeds 
$N_q$, visual tokens are the primary driver of overhead. 
Visual token compression therefore targets the prefill 
bottleneck by shortening the multimodal sequence before or 
during LLM processing.

\subsection{Token Compression for Video-LLMs}

\paragraph{Training-free visual token compression.}
Training-free token compression for Video-LLMs has been 
explored along several directions. Among outer-LLM methods, spatial approaches 
reduce within-frame redundancy using per-frame signals 
such as attention-based saliency, token diversity, coverage, or 
information uniqueness~\cite{visionzip, unicomp, divprune, nuwa}. 
Spatiotemporal methods address both sources of redundancy 
through diverse strategies. Some explicitly partition 
videos at scene transitions and compress within each 
segment~\cite{fastvid, flashvid}. Others constrain 
temporal matching to spatially nearby tokens across 
adjacent frames, for example by dynamic programming over 
temporal windows~\cite{holitom} or connecting tokens into 
forests under spatiotemporal distance 
constraints~\cite{forestprune}. A further line of work 
bypasses locality altogether and selects tokens from the 
full spatiotemporal pool, either through submodular 
optimization~\cite{floc} or unified global 
selection~\cite{unistc}. A concurrent line of work also explores optimal transport for video token compression. AOT~\cite{aot} applies OT to aggregate token information into pre-selected anchors at 
both spatial and temporal levels, but leaves the 
compression decision itself outside the transport 
formulation. In contrast, our work treats temporal 
compression as a transport-based allocation problem, where 
importance-aware mass and pairwise transport difficulty 
jointly determine token-level compression and budget 
distribution across frame pairs.
\section{Method}

\begin{figure*}[t]
\centering
\includegraphics[width=\textwidth]{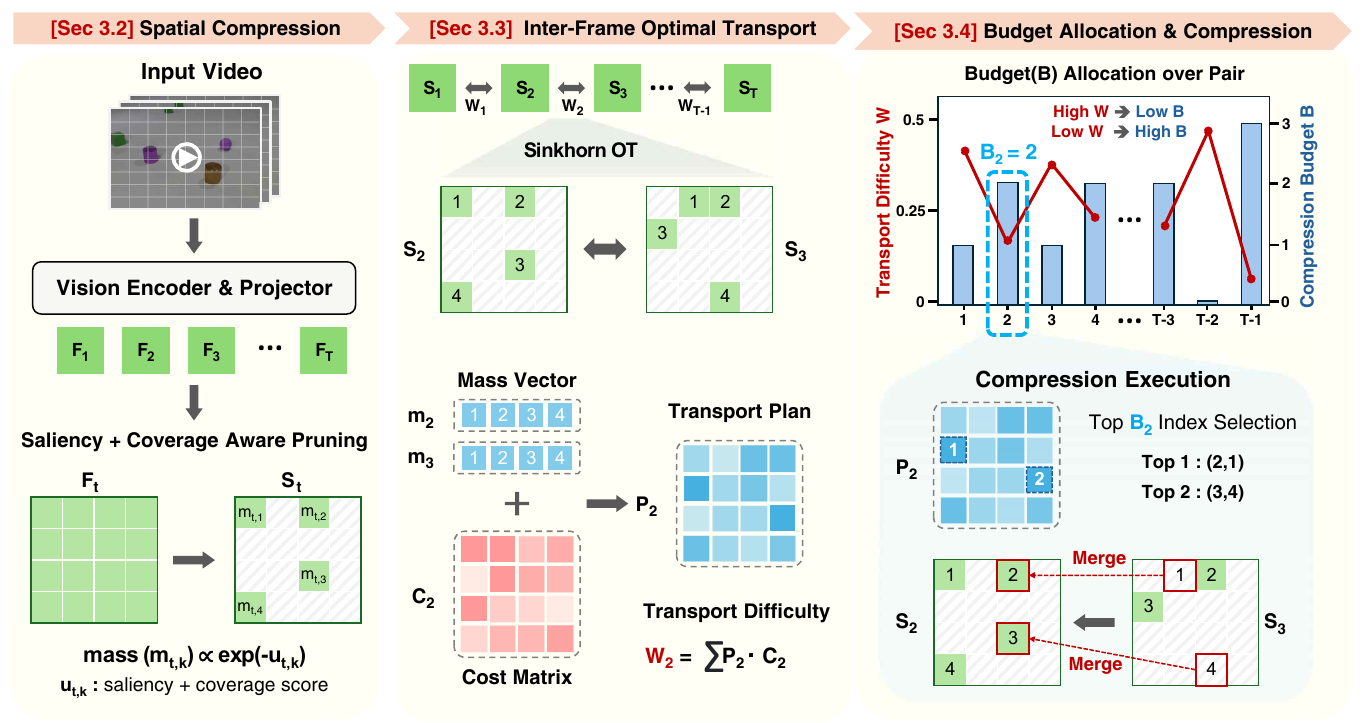}
\caption{
Overview of \textbf{OTT-Vid}. \textbf{(Sec~\ref{sec:spatial})} Each frame $t$ is encoded into visual tokens $F_t$, spatially pruned to retained tokens $S_t$, and converted into a mass vector $m_t$ from per-token importance. \textbf{(Sec~\ref{sec:temporal})} Sinkhorn OT computes a transport plan $P_t$ and transport difficulty $W_t$ for each neighboring frame pair. \textbf{(Sec~\ref{sec:budget})} Compression budgets $B_t$ are allocated in inverse proportion to transport difficulty, and the transport plan determines which tokens to compress.
}
\vspace{-0.4cm}
\label{fig:overview}
\end{figure*}

\subsection{Overview}
\label{sec:overview}
An overview of \textbf{OTT-Vid} is provided in 
Figure~\ref{fig:overview}. A video of $T$ frames is 
encoded into $N_v$ visual tokens per frame by the vision 
encoder and projector, yielding $T \cdot N_v$ tokens in 
total. Given a target retention ratio $r$, OTT-Vid compresses this set through two stages, governed by a temporal share parameter $\gamma \in [0,1]$ that 
controls the balance between spatial and temporal 
compression. The total retention ratio $r$ is decomposed 
as $r = r_s \cdot r_t$, where $r_s = r^{1-\gamma}$ and 
$r_t = r^{\gamma}$. In 
\textbf{Sec.~\ref{sec:spatial}}, we first prune each 
frame independently to remove spatial redundancy, 
retaining $K = N_v \cdot r_s$ tokens per frame and 
constructing a non-uniform mass for each retained token 
according to its frame-wise importance. In the temporal 
stage, we solve optimal transport between neighboring 
frame pairs to obtain two decision signals: a transport 
coupling that ranks compression candidates under 
importance-aware mass and matching cost 
(\textbf{Sec.~\ref{sec:temporal}}), and a transport 
difficulty that estimates pairwise compressibility. We then distribute a total budget of $B_{\mathrm{tot}} = K \cdot T \cdot (1 - r_t)$ compression operations across all $T{-}1$ adjacent frame pairs 
according to this difficulty, and simultaneously execute the allocated compression on all pairs in parallel (\textbf{Sec.~\ref{sec:budget}}).

\subsection{Spatial Compression and Mass Assignment}
\label{sec:spatial}
The first stage of our framework compresses each frame independently and
assigns a transport mass to each retained token. Prior spatial token
selection methods \cite{scope,flashvid} have explored semantic importance and representativeness
as complementary criteria: importance preserves informative content, while
representativeness prevents the selected tokens from collapsing onto a few
redundant regions. We instantiate this principle with a
saliency-weighted coverage objective, where coverage is weighted by the
saliency of the tokens being covered, and pass the selected tokens to the
temporal OT stage.
\paragraph{Saliency-weighted representative token selection.}
For each frame $t$ with $N_v$ tokens $\{x_{t,i}\}_{i=1}^{N_v}$ 
where $x_{t,i} \in \mathbb{R}^{d}$, our goal is to retain 
$K$ tokens. A saliency score $w_{t,i}$ is computed from the last layer 
of the vision encoder by averaging its self-attention 
weights over all heads and query positions, normalized 
such that $\sum_i w_{t,i} = 1$. We then greedily construct 
a representative subset $\mathcal{S}_t$ of size $K$ by 
maximizing saliency-weighted coverage. Starting from an empty set, 
we iteratively add the token whose marginal gain is largest. 
The gain of a candidate token $j$ measures how much 
additional salient content it would cover beyond what the 
current selected set already explains:
\begin{equation}
\mathrm{gain}(j)
=
\sum_{i=1}^{N_v}
w_{t,i}
\max\bigl(0,\,
\mathrm{sim}(x_{t,i},\,x_{t,j}) - \mu_i
\bigr),
\end{equation}
where $\mathrm{sim}(\cdot,\cdot)$ denotes cosine similarity 
and $\mu_i = \max_{v \in \mathcal{S}_t} 
\mathrm{sim}(x_{t,i}, v)$ measures how well token 
$x_{t,i}$ is currently covered by the selected set, 
initialized to zero. The weighting by $w_{t,i}$ ensures 
that improvements in covering more salient tokens 
contribute more to the gain. We update $\mu_i$ after each 
addition and repeat until $K$ tokens are retained.

\paragraph{Leave-one-out mass assignment.}
To quantify the preservation priority of each selected 
token $s_{t,k} \in \mathcal{S}_t$, we estimate its 
irreplaceable semantic contribution. Concretely, we ask: 
if $s_{t,k}$ were removed from the selected set, how 
much representation quality would the original frame 
lose? We first induce a nearest-neighbor partition of the original
tokens by assigning each $x_{t,i}$ to its best-matching retained
token $\hat{k}(i)$ in $\mathcal{S}_t$. Let $\sigma_{i,1}$ and $\sigma_{i,2}$ denote the largest and second-largest cosine similarities between $x_{t,i}$ and tokens in $\mathcal{S}_t$. If the best match $s_{t,\hat{k}(i)}$ were removed, the second-best would 
take over, and the gap $\sigma_{i,1} - \sigma_{i,2}$ 
quantifies the resulting loss in representation quality.
We define the contribution $u_{t,k}$ by aggregating these 
saliency-weighted gaps over the original tokens for which 
$s_{t,k}$ is the best match, and convert it into a 
mass $m_{t,k}$ via a negative softmax with temperature $\tau_m$:
\begin{equation}
u_{t,k} = \sum_{i:\hat{k}(i)=k} 
w_{t,i} \max\bigl(0,\, \sigma_{i,1} - \sigma_{i,2}\bigr),
\quad
\hat{k}(i) = \arg\max_{k}\, \mathrm{sim}(x_{t,i},\, s_{t,k}),
\end{equation}
\begin{equation}
\label{eq:mass_assignment}
\tilde{u}_{t,k} 
= u_{t,k} \big/ \max_{k'} u_{t,k'},
\qquad
m_{t,k} 
= \exp(-\tilde{u}_{t,k}/\tau_m) \big/ \sum_{k'} \exp(-\tilde{u}_{t,k'}/\tau_m).
\end{equation}
The negation inverts priority so that important tokens receive smaller mass and resist compression, 
while the softmax normalizes total mass to one for 
consistent scaling across frame pairs.

\subsection{Temporal Compression via Optimal Transport}
\label{sec:temporal}
Let $\mathcal{S}_t=\{s_{t,i}\}_{i=1}^{K}$ and
$\mathcal{S}_{t+1}=\{s_{t+1,j}\}_{j=1}^{K}$ denote the
retained token sets of two neighboring frames, with
corresponding mass vectors $m_t, m_{t+1} \in
\mathbb{R}_{+}^{K}$ constructed in 
Sec.~\ref{sec:spatial}. We formulate temporal compression as an optimal transport problem balancing importance-aware mass and matching cost. Since OT's marginal constraint prevents small-mass tokens from accumulating large coupling entries, our mass construction directly encodes preservation priority into the transport plan, with larger coupling values identifying high-priority compression candidates.

\paragraph{Transport cost.}
The cost matrix $C_t \in \mathbb{R}^{K \times K}$ assigns a transport penalty to each cross-frame token pair, with smaller values indicating preferred correspondences. Existing methods either define this cost purely by semantic similarity~\cite{flashvid,fastvid}, which risks pairing visually similar tokens from different objects, or restrict matching to identical grid positions~\cite{sttm,holitom}, which leaves few valid matches under camera panning or substantial object motion. To benefit from both while mitigating their weaknesses, we combine a semantic dissimilarity term with a spatial distance term and adapt their relative weight according to the motion characteristics of each frame pair. We define semantic dissimilarity by cosine distance and 
spatial dissimilarity by scaled Euclidean distance on 
the patch grid, where $p_{t,i}$ and $p_{t+1,j}$ are 
the patch-grid positions and $d_{\max}$ is the grid 
diagonal length. The two terms are combined with a 
mixing coefficient $\alpha_t$:
\begin{equation}
c_{\mathrm{sem}}(i,j)
= 1 - \mathrm{sim}(s_{t,i},\, s_{t+1,j}),
\qquad
c_{\mathrm{loc}}(i,j) = \|p_{t,i} - p_{t+1,j}\|_2 \big/ d_{\max},
\end{equation}
\begin{equation}
C_{t,ij}
= \alpha_t\,c_{\mathrm{sem}}(i,j)
+ (1 - \alpha_t)\,c_{\mathrm{loc}}(i,j).
\label{eq:cost_matrix}
\end{equation}
The mixing coefficient $\alpha_t$ is computed as
$\alpha_t = 1 - \bar{s}_t/2$, where $\bar{s}_t$ is the mean cosine
similarity between co-located token pairs in the original
(pre-pruning) frames $t$ and $t{+}1$, clamped to $[0,1]$. We use
$\bar{s}_t$ as a lightweight proxy for scene dynamics. When $\bar{s}_t$ is high, the frame pair is likely to be mostly static 
and co-located patches provide reliable correspondence cues, so we 
give locality a stronger role to discourage matches between visually 
similar but spatially distant regions. Conversely, low $\bar{s}_t$
suggests camera motion, object motion, or a scene transition, where
fixed spatial positions are less reliable. In this case, semantic
dissimilarity remains more influential, allowing correspondences across
larger spatial displacements. This adaptive behavior is bounded by the $1/2$ scaling, which confines $\alpha_t$ to $[1/2, 1]$ so that semantic dissimilarity remains the dominant cost component while locality can contribute up to an equal weight in highly static frame pairs.

\paragraph{Transport plan and compression difficulty.}
Given the mass vectors and the cost matrix, we solve an
entropically regularized OT problem via Sinkhorn iterations~\cite{sinkhorn}:
\begin{equation}
P_t
= \operatorname*{argmin}_{P \in \Pi(m_t, m_{t+1})}
\sum_{i,j} P_{ij}\, C_{t,ij} - \varepsilon H(P),
\label{eq:ot_plan}
\end{equation}
where $\Pi(m_t, m_{t+1})$ denotes all valid 
transport plans whose rows sum to $m_t$ and columns 
sum to $m_{t+1}$, and $H(P) = -\sum_{i,j} P_{ij} 
\log P_{ij}$ is the entropic regularizer that smooths 
the solution.

The resulting transport plan $P_t$ specifies an optimal coupling, where larger $P_{t,ij}$ entries indicate high-priority candidates for compression. Beyond individual coupling values, we also quantify the overall transport difficulty of each frame pair as the total cost incurred by the transport plan:
\begin{equation}
W_t = \sum_{i,j} P_{t,ij}\, C_{t,ij}.
\end{equation}

\subsection{Transport-Derived Budget Allocation}
\label{sec:budget}

\paragraph{Per-pair budget allocation.}
The transport difficulty $W_t$ quantifies how easily 
the retained tokens in neighboring frame pair 
$(t, t{+}1)$ can be matched. Since each cost is 
weighted by its transport mass, $W_t$ primarily 
reflects the matching cost of redundant tokens, while 
important tokens (with smaller mass) contribute little. 
A smaller $W_t$ thus indicates that redundant tokens 
find low-cost counterparts, implying the pair is more 
compressible. We therefore use 
$\{W_t\}_{t=1}^{T-1}$ to allocate the temporal 
compression budget non-uniformly via a negative softmax 
with temperature $\tau_b$, so that pairs with smaller 
$W_t$ receive larger budgets:
\begin{equation}
\beta_t
=
\exp(-W_t / \tau_b)
\Big/
\sum_{t'=1}^{T-1}\exp(-W_{t'} / \tau_b),
\qquad
B_t = \mathrm{round}(\beta_t B_{\mathrm{tot}}),
\label{eq:budget_allocation}
\end{equation}

\paragraph{Compression execution.}
Given each transport plan $P_t$ and its allocated budget $B_t$, we 
select compression candidates from frame $t{+}1$ (source) to frame 
$t$ (destination) by ranking pairs according to their coupling entries $P_{t,ij}$ and taking the top-$B_t$ entries, under the constraint that each source token is selected at most once. This yields a per-pair many-to-one assignment, in which multiple source tokens may map to the same destination while unselected source tokens are retained. The top-$B_t$ selection may include pairs with high transport cost when low-cost correspondences are scarce. Since high cost indicates a semantically unreliable match, we merge a selected pair only when $C_{t,ij} < \tau_c$, otherwise pruning the source token to avoid contaminating the merged representation.

All $T{-}1$ transport plans are computed from the initial 
spatially-pruned sets $\{S_t\}$, and the resulting matches form edges in a global directed graph. Since a token may serve as both destination and source across consecutive pairs, we resolve the resulting chains via union-find, with each connected component collapsed to its root by uniform averaging. This generalizes per-pair merging to the full sequence in a single parallel pass.

\section{Experimental Results}

\paragraph{Benchmarks.}
We evaluate on four video question answering (VQA) benchmarks: 
MVBench~\cite{mvbench}, VideoMME~\cite{videomme}, 
LongVideoBench~\cite{longvideobench}, and MLVU~\cite{mlvu}, 
and on two video temporal grounding (VTG) benchmarks. For VTG, we use the recently refined Charades-TimeLens and ActivityNet-TimeLens 
splits~\cite{timelens}, which provide manually 
re-annotated versions of Charades-STA~\cite{charades} 
and ActivityNet Captions~\cite{activitynet} with 
improved annotation quality. VTG requires 
precise event localization and is therefore particularly 
sensitive to the preservation of temporal evidence 
during compression.

\paragraph{Implementation details.}
The default hyperparameters in our experiments are 
$\gamma = \tau_m = \tau_b = \tau_c = 0.3$ for the temporal 
share, mass temperature, budget temperature, and cost 
threshold, respectively. Evaluations are performed on three 
Video-LLMs: Qwen2.5-VL-7B~\cite{qwen2.5vl} (32 frames for 
VQA, 2 fps for temporal grounding with dynamic resolution), 
LLaVA-OneVision-7B~\cite{llava-ov} (32 frames, 196 tokens 
per frame), and LLaVA-Video-7B~\cite{llava-video} (64 frames, 
169 tokens per frame). Additional implementation details are 
provided in Appendix~\ref{appendix:implementation}.

\paragraph{Compared methods.}
We compare \textbf{OTT-Vid} against four state-of-the-art 
training-free video token compression methods. 
\textbf{HoliTom}~\cite{holitom} performs redundancy-aware temporal 
segmentation via dynamic programming, followed by 
spatiotemporal token merging. \textbf{FastVID}~\cite{fastvid} 
segments videos at scene transitions and applies 
density-based spatiotemporal pruning within each segment. 
\textbf{FlashVID}~\cite{flashvid} selects representative tokens via 
saliency and diversity, then merges remaining tokens 
through tree-structured spatiotemporal compression. 
\textbf{UniComp}~\cite{unicomp} introduces information uniqueness 
as a unified criterion for frame-level fusion, per-frame 
token allocation, and spatial compression. Since \textbf{HoliTom}~\cite{holitom} 
and \textbf{FlashVID}~\cite{flashvid} include inner-LLM compression stages, we 
disable these components and retain only their outer-LLM 
compression to ensure fair comparison at the same 
retention ratio.

\subsection{Comparisons with State-of-the-Art Methods}
\label{sec:main_results}

\paragraph{Main results on Qwen2.5-VL.}
Table~\ref{tab:main} compares OTT-Vid with recent training-free token compression methods on Qwen2.5-VL-7B across four VQA and two VTG benchmarks. OTT-Vid achieves the best average retention at every token budget on both task families, and the margin grows as compression becomes more aggressive. At the most challenging 10\% retention, OTT-Vid preserves 95.8\% of the uncompressed VQA performance and 73.9\% of VTG performance, outperforming the strongest competing baseline by \textbf{1.0} and \textbf{7.3} percentage points, respectively.

The advantage is particularly pronounced on temporal grounding, which directly probes whether compression preserves fine-grained temporal structure. Existing similarity-driven methods~\cite{flashvid,fastvid} exhibit substantial VTG degradation under tight budgets, dropping to 51.3--66.6\% retention at 10\%, while OTT-Vid maintains 73.9\%. The gap widens substantially as the budget shrinks. This consistent VTG advantage supports our central claim that incorporating intra-frame token importance and transport-derived budget allocation is critical for handling redundancy in the temporal domain.

\paragraph{Generalization across Video-LLM backbones.}
Table~\ref{tab:cross_backbone} reports VQA results on LLaVA-OneVision-7B and LLaVA-Video-7B (these backbones do not support temporal grounding). OTT-Vid remains competitive across both backbones at every retention ratio, ranking first or within 0.6 points of the best baseline. On LLaVA-Video-7B at 10\% retention, OTT-Vid achieves the highest average retention of 94.5\%. These results indicate that the benefits of importance-aware mass and transport-derived budget allocation are not tied to a specific Video-LLM and transfer across backbones with different frame counts and per-frame token configurations.

\begin{table*}[t]
\centering
\caption{Main results on Qwen2.5-VL-7B across four VQA 
and two VTG benchmarks. VQA scores are reported as 
accuracy and VTG scores as mIoU. Avg. (\%) denotes the 
mean performance retention relative to the uncompressed 
baseline. Best results per retention ratio are in 
\textbf{bold}.}
\label{tab:main}
\resizebox{\textwidth}{!}{
\setlength{\tabcolsep}{4pt}
\renewcommand{\arraystretch}{0.85}
\begin{tabular}{l | c | cccc | c || cc | c}
\toprule
& & \multicolumn{5}{c||}{\textbf{Video Question Answering}} 
& \multicolumn{3}{c}{\textbf{Video Temporal Grounding}} \\
\cmidrule(lr){3-7} \cmidrule(lr){8-10}
\multirow{-2}{*}{\textbf{Method}} 
& \multirow{-2}{*} {\textbf{Ret.}}
& MVBench & VideoMME & LVB & MLVU & Avg. (\%)
& Charades & ANet & Avg. (\%) \\
\midrule
Vanilla & 100\% 
& 67.8 & 62.1 & 58.8 & 62.1 & 100.0
& 36.7 & 30.7 & 100.0 \\
\midrule
HoliTom~\cite{holitom} {\scriptsize NeurIPS'25} & 25\% 
& 66.1 & 60.3 & 57.8 & 61.3 & 97.9
& 35.2 & 24.3 & 87.5 \\
FastVID~\cite{fastvid} {\scriptsize NeurIPS'25} & 25\% 
& 65.1 & 59.0 & 57.5 & 61.1 & 96.8
& 25.5 & 20.8 & 68.5 \\
FlashVID~\cite{flashvid} {\scriptsize ICLR'26} & 25\% 
& 66.0 & 58.7 & \textbf{58.5} & 61.6 & 97.6
& 33.1 & 16.6 & 72.2 \\
UniComp~\cite{unicomp} {\scriptsize CVPR'26} & 25\% 
& 65.4 & \textbf{60.6} & 58.1 & 61.6 & 98.0
& 35.0 & 24.5 & 87.5 \\
\rowcolor{bestrow}
\textbf{OTT-Vid} & 25\% 
& \textbf{66.5} & 60.4 
& 58.3 & \textbf{61.8} & \textbf{98.5}
& \textbf{35.9} & \textbf{25.5} & \textbf{90.3} \\
\midrule
HoliTom~\cite{holitom} {\scriptsize NeurIPS'25} & 20\% 
& 66.1 & 60.0 & 57.1 & \textbf{61.6} & 97.6
& 34.3 & 22.3 & 82.9 \\
FastVID~\cite{fastvid} {\scriptsize NeurIPS'25} & 20\% 
& 65.1 & 58.2 & 57.2 & 60.8 & 96.2
& 25.2 & 19.9 & 66.5 \\
FlashVID~\cite{flashvid} {\scriptsize ICLR'26} & 20\% 
& 65.8 & 57.6 & \textbf{57.8} & 60.8 & 96.5
& 31.9 & 14.2 & 66.5 \\
UniComp~\cite{unicomp} {\scriptsize CVPR'26} & 20\% 
& 64.8 & \textbf{61.0} & 57.4 & 61.3 & 97.5
& 34.6 & 22.7 & 84.0 \\
\rowcolor{bestrow}
\textbf{OTT-Vid} & 20\% 
& \textbf{66.3} & \textbf{61.0} 
& \textbf{57.8} & 61.5 & \textbf{98.2}
& \textbf{34.8} & \textbf{24.0} & \textbf{86.5} \\
\midrule
HoliTom~\cite{holitom} {\scriptsize NeurIPS'25} & 15\% 
& 65.4 & 59.2 & 56.8 & 61.0 & 96.6
& 32.5 & 19.5 & 76.0 \\
FastVID~\cite{fastvid} {\scriptsize NeurIPS'25} & 15\% 
& 64.2 & 58.2 & 57.2 & 59.9 & 95.5
& 24.8 & 18.1 & 63.2 \\
FlashVID~\cite{flashvid} {\scriptsize ICLR'26} & 15\% 
& 65.1 & 56.9 & 57.1 & 61.3 & 95.8
& 29.8 & 11.5 & 59.2 \\
UniComp~\cite{unicomp} {\scriptsize CVPR'26} & 15\% 
& 64.1 & 59.8 & 56.4 & 59.5 & 95.6
& 33.2 & 20.2 & 78.0 \\
\rowcolor{bestrow}
\textbf{OTT-Vid} & 15\% 
& \textbf{65.5} & \textbf{60.4} 
& \textbf{57.5} & \textbf{61.4} & \textbf{97.6}
& \textbf{33.7} & \textbf{21.8} & \textbf{81.5} \\
\midrule
HoliTom~\cite{holitom} {\scriptsize NeurIPS'25} & 10\% 
& \textbf{64.2} & 57.9 & 55.6 & 60.1 & 94.8
& 28.0 & 14.8 & 62.3 \\
FastVID~\cite{fastvid} {\scriptsize NeurIPS'25} & 10\% 
& 63.2 & 55.9 & 55.6 & 59.6 & 93.4
& 23.9 & 15.7 & 58.0 \\
FlashVID~\cite{flashvid} {\scriptsize ICLR'26} & 10\% 
& 63.7 & 55.2 & \textbf{56.9} & 59.7 & 93.9
& 26.1 & 9.8 & 51.3 \\
UniComp~\cite{unicomp} {\scriptsize CVPR'26} & 10\% 
& 61.9 & 58.2 & 54.2 & 57.4 & 92.4
& 29.7 & 16.0 & 66.6 \\
\rowcolor{bestrow}
\textbf{OTT-Vid} & 10\% 
& 64.0 & \textbf{58.7} 
& 56.7 & \textbf{60.7} & \textbf{95.8}
& \textbf{32.3} & \textbf{18.4} & \textbf{73.9} \\
\bottomrule
\end{tabular}
}
\end{table*}

\begin{table*}[t]
\centering
\caption{Cross-backbone evaluation on LLaVA-OneVision-7B and LLaVA-Video-7B (VQA only, as these models do not support temporal grounding). Avg.~(\%) denotes the mean performance retention relative to the uncompressed baseline. Best results per setting are in \textbf{bold}.}
\label{tab:cross_backbone}
\footnotesize
\setlength{\tabcolsep}{6pt}
\renewcommand{\arraystretch}{0.85}
\resizebox{\textwidth}{!}{
\begin{tabular}{l|c|cccc|c||cccc|c}
\toprule
& & \multicolumn{5}{c|}{\textbf{LLaVA-OV-7B}} & \multicolumn{5}{c}{\textbf{LLaVA-Video-7B}} \\
\cmidrule(lr){3-7} \cmidrule(lr){8-12}
\multirow{-2}{*}{\textbf{Method}} 
& \multirow{-2}{*} {\textbf{Ret.}} & MVB & VMME & LVB & MLVU & Avg(\%) & MVB & VMME & LVB & MLVU & Avg(\%) \\
\midrule
Vanilla   & 100\% & 58.4 & 58.6 & 56.4 & 63.3 & 100.0 & 62.3 & 64.4 & 59.9 & 68.2 & 100.0 \\
\midrule
HoliTom   & 25\%  & \textbf{58.5} & \textbf{58.9} & 57.1 & 63.1 & \textbf{100.4} & 60.9 & 62.7 & 57.8 & 65.9 & 97.0 \\
FastVID   & 25\%  & 58.3 & 58.0 & 55.8 & 61.4 & 98.7  & \textbf{61.1} & \textbf{63.8} & 57.7 & 66.3 & \textbf{97.6} \\
FlashVID  & 25\%  & 58.4 & 58.2 & 57.1 & 63.4 & 100.2 & 59.9 & 61.7 & \textbf{58.8} & 66.3 & 96.8 \\
UniComp   & 25\%  & 57.7 & \textbf{58.9} & \textbf{57.7} & 62.4 & 100.0 & 58.3 & 60.8 & 58.3 & 65.1 & 95.1 \\
\rowcolor{bestrow}
\textbf{OTT-Vid} & 25\%  & 58.0 & 58.2 & 56.9 & \textbf{63.5} & 100.0  & 60.8 & 62.2 & 58.3 & \textbf{66.7} & 97.5 \\
\midrule
HoliTom   & 10\%  & \textbf{57.0} & 57.5 & 56.2 & 60.4 & 97.7  & \textbf{59.8} & 61.2 & 55.9 & 62.9 & 94.1 \\
FastVID   & 10\%  & \textbf{57.0} & 56.8 & 55.1 & 60.3 & 96.8  & 59.7 & 60.2 & 56.8 & 63.3 & 94.2 \\
FlashVID  & 10\%  & 56.3 & 56.7 & 54.2 & \textbf{61.9} & 96.8  & 59.3 & 59.9 & 56.4 & 63.4 & 93.8 \\
UniComp   & 10\%  & 56.4 & \textbf{58.0} & \textbf{57.4} & 60.9 & \textbf{98.4} & 55.9 & 59.3 & \textbf{57.0} & \textbf{64.8} & 93.0 \\
\rowcolor{bestrow}
\textbf{OTT-Vid} & 10\%  & 56.6 & 57.1 & 56.1 & 61.5 & 97.8  & 59.2 & \textbf{61.5} & 55.8 & 64.4 & \textbf{94.5} \\
\bottomrule
\end{tabular}
}
\end{table*}

\begin{figure*}[t]
\centering
\begin{minipage}{0.48\textwidth}
\centering
\captionof{table}{Ablation on the core components of 
OTT-Vid at 10\% retention on Qwen2.5-VL-7B. Mass denotes 
importance-aware mass (\checkmark) vs.\ uniform. Budget 
denotes adaptive allocation (\checkmark) vs.\ uniform.}
\label{tab:ablation}
\setlength{\tabcolsep}{4pt}
\small
\begin{tabular}{c cc cc}
\toprule
\multirow{2}{*}{\textbf{Row}} & \multicolumn{2}{c}{\textbf{Method}} 
& \multicolumn{2}{c}{\textbf{Performance (\%)}} \\
\cmidrule(lr){2-3} \cmidrule(lr){4-5}
& Mass & Budget & VQA Avg. & VTG Avg. \\
\midrule
(a) & & & 94.3 & 63.1 \\
(b) & & \checkmark & 94.6 & 64.3 \\
(c) & \checkmark & & 95.2 & 72.2 \\
\rowcolor{bestrow}
(d) & \checkmark & \checkmark & \textbf{95.8} & \textbf{73.9} \\
\bottomrule
\end{tabular}
\end{minipage}
\hfill
\begin{minipage}{0.48\textwidth}
\centering
\captionof{figure}{Ablation on temporal share $\gamma$ on average performance and compression latency at 10\% retention. Average performance is the mean of VQA and VTG retention.}
\includegraphics[width=\textwidth]{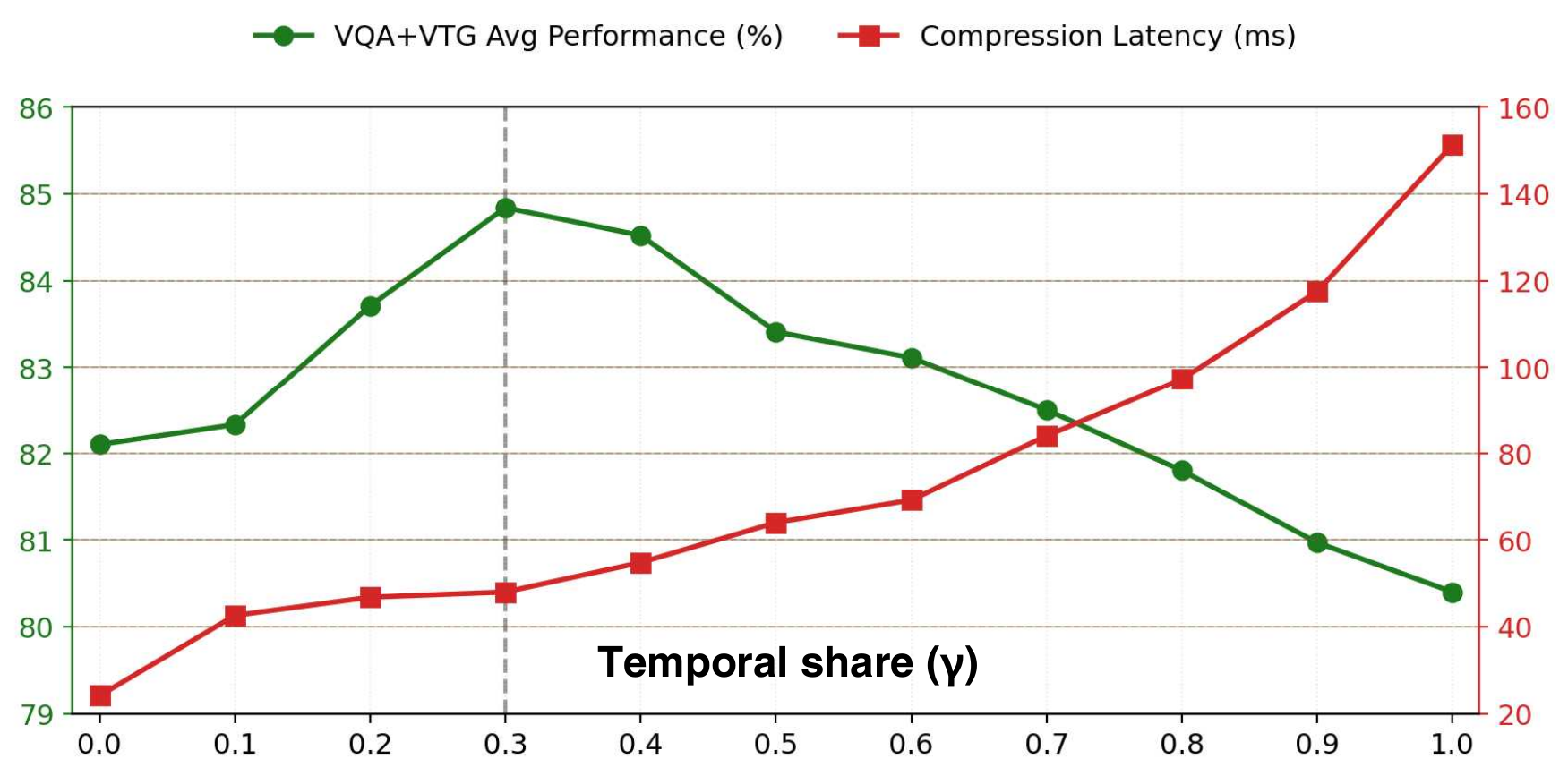}

\label{fig:ts_ablation}
\end{minipage}
\end{figure*}
\begin{table}[t]
\centering
\caption{Efficiency and performance at 10\% retention on 
Qwen2.5-VL-7B. VQA and VTG report mean performance 
retention (\%) relative to the uncompressed baseline.}
\label{tab:efficiency}
\resizebox{\columnwidth}{!}{
\setlength{\tabcolsep}{7pt}
\renewcommand{\arraystretch}{0.95}
\begin{tabular}{l | c | cccc | c | c | cc}
\toprule
& & \multicolumn{4}{c|}{\textbf{Latency (ms) }} 
& & & \multicolumn{2}{c}{\textbf{Perf. (\%) }} \\
\cmidrule(lr){3-6} \cmidrule(lr){9-10}
\multirow{-2}{*}{\textbf{Method}} 
& \multirow{-2}{*}{\textbf{Ret.}}
& Vision & Comp. & LLM & TTFT 
& \multirow{-2}{*}{\textbf{Peak Mem. (GB) }} 
& \multirow{-2}{*}{\textbf{TFLOPs }} 
& VQA & VTG \\
\midrule
Vanilla & 100\%
& 786.0 & -- & 1088.8 & 1874.8 (1.0$\times$) 
& 17.76 & 120.25 
& 100.0 & 100.0 \\
FastVID & 10\%
& 786.0 & 17.3 & 104.0 & 907.3 (2.1$\times$) 
& 17.90 & 26.11 
& 93.4 & 58.0 \\
UniComp & 10\%
& 786.0 & 286.1 & 104.0 & 1176.1 (1.6$\times$) 
& 23.17 & 26.07 
& 92.4 & 66.6 \\
\rowcolor{bestrow}
\textbf{OTT-Vid} & 10\%
& 786.0 & 48.0 & 104.0 & 938.0 (2.0$\times$) 
& 18.22 & 26.11 
& 95.8 & 73.9 \\
\bottomrule
\end{tabular}
}
\end{table}

\begin{figure}[t]
\centering
\includegraphics[width=\columnwidth, trim=0 0 0 0, clip]{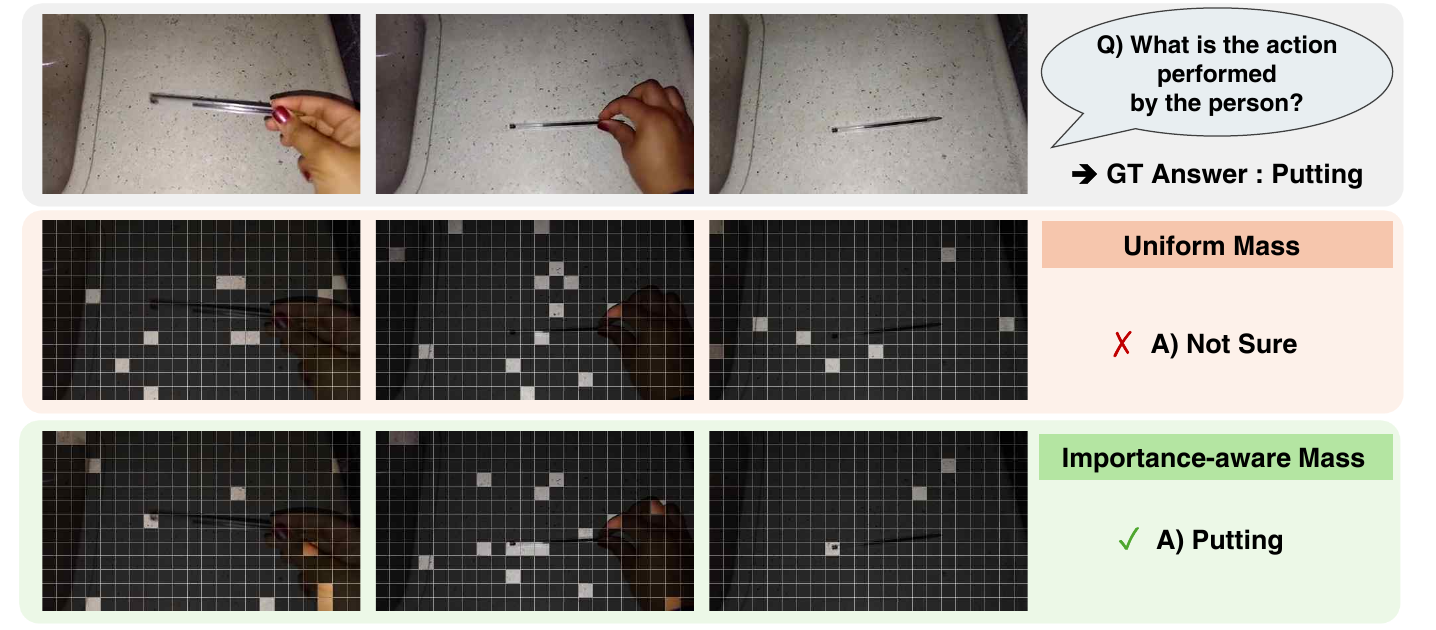}
\caption{Visualization result of token compression 
with uniform vs.\ importance-aware mass at 10\% retention. Bright patches indicate retained tokens. Uniform mass loses tokens around the 
hand and object, while importance-aware mass preserves 
these semantically critical regions.}
\label{fig:visualize}
\end{figure}

\subsection{Ablation Studies}

\paragraph{Ablation on core components.}
Table~\ref{tab:ablation} ablates the two core components 
of OTT-Vid at 10\% retention. Row (a) uses uniform mass 
and uniform budget as the non-adaptive baseline, while 
rows (b) and (c) enable each component independently. 
Mass alone (c) yields a substantial gain over (a), 
particularly on VTG (+9.1 percentage points), supporting our claim that 
jointly considering token importance and similarity helps 
preserve temporal evidence. Budget alone (b) provides a smaller but positive improvement, 
more pronounced on benchmarks with greater variation in temporal compressibility (refer to Table~\ref{tab:ablation_per_benchmark} in Appendix).
Their combination (d) achieves the best results on both 
task families, indicating that the two components address 
complementary sources of redundancy.

\paragraph{Ablation on temporal share \texorpdfstring{$\gamma$}{gamma}.}
The temporal share $\gamma$ controls the balance between spatial pruning and temporal compression. Since the per-frame token count $K = N_v \cdot r_s$ governs the cost of greedy selection, leave-one-out mass computation, and the OT solver, latency grows with $\gamma$. Figure~\ref{fig:ts_ablation} reports the average performance retention across all six benchmarks (the mean of VQA and VTG averages) against compression latency (ms) as $\gamma$ varies at 10\% retention. Average retention peaks at $\gamma = 0.3$ with 84.9\% and degrades toward both extremes, falling to 82.1\% at $\gamma = 0$ and 80.4\% at $\gamma = 1$, while latency rises from 24.1 ms at $\gamma = 0$ to 151.3 ms at $\gamma = 1$. The setting $\gamma = 0.3$ thus offers the best trade-off and is set as the default.

Additional ablations on $\tau_b$, $\tau_m$, $\tau_c$, 
and the mixing coefficient $\alpha_t$ are provided in 
Appendix~\ref{appendix:ablation}.

\subsection{Efficiency Analysis}
Table~\ref{tab:efficiency} compares OTT-Vid with FastVID~\cite{fastvid} and UniComp~\cite{unicomp} at $r{=}10\%$ on Qwen2.5-VL-7B. OTT-Vid achieves a $2.0\times$ TTFT speedup with 18.22\,GB peak memory, comparable to FastVID ($2.1\times$, 17.90\,GB) and improving over UniComp ($1.6\times$, 23.17\,GB). All three methods reduce TFLOPs to roughly 26 from the uncompressed 120.25, indicating that the dominant inference cost is governed by the retention ratio rather than the compression algorithm. Under this comparable efficiency profile, OTT-Vid maintains the strongest accuracy across both VQA and VTG benchmarks.

\subsection{Qualitative Analysis}
Figure~\ref{fig:visualize} compares retained tokens under 
uniform mass and importance-aware mass at 10\% retention 
on a video of a person putting down a pen. With uniform 
mass, tokens are distributed regardless of semantic role, 
and the regions corresponding to the hand and the pen are 
largely discarded, leading the model to fail at identifying 
the action. In contrast, importance-aware mass concentrates 
the budget on tokens carrying the action evidence, 
preserving the hand and the pen across frames and yielding 
the correct prediction. This illustrates how encoding token 
preservation priority into the transport constraints 
translates OTT-Vid's quantitative gains into more 
semantically faithful compression. Additional qualitative 
results are in Appendix~\ref{appendix:visualize}.

\section{Conclusion}
We presented OTT-Vid, a training-free video token compression 
framework that formulates temporal compression as an optimal 
transport problem with adaptive budget allocation across 
frame pairs. Across six benchmarks, OTT-Vid retains 95.8\% 
VQA and 73.9\% VTG performance at 10\% retention, 
outperforming existing training-free baselines. The advantage 
is most pronounced on temporal grounding, where prior 
similarity-driven methods degrade sharply.

\paragraph{Limitations and Future Work.}
OTT-Vid computes all pairwise transport plans in parallel 
from the initial spatially-pruned representations, which 
enables single-pass global budget allocation with low 
overhead but does not account for representational changes 
induced by compression on earlier pairs. Sequential 
execution that propagates these changes could refine 
transport plans at the cost of higher computational overhead, 
which we leave for future work.

\FloatBarrier
\newpage

\bibliographystyle{plainnat}
\bibliography{reference}

\newpage

\appendix

\setcounter{table}{0}
\setcounter{figure}{0}
\setcounter{equation}{0}

\renewcommand{\thetable}{A\arabic{table}}
\renewcommand{\thefigure}{A\arabic{figure}}
\renewcommand{\theequation}{A\arabic{equation}}

\section{Additional Implementation Details}
\label{appendix:implementation}

\paragraph{Sinkhorn parameters.}
The OT transport plan in Eq.~\eqref{eq:ot_plan} is 
solved using the Sinkhorn algorithm with entropic 
regularization parameter $\epsilon{=}0.01$ and a maximum of 
200 iterations. We choose a small $\epsilon$ since our 
compression executes discrete decisions: each source 
token is either fully assigned to a target token (merge) or 
removed (prune), without fractional mass splitting. A small 
$\epsilon$ produces a sharper transport plan that more 
closely approximates the discrete optimal assignment, while 
larger values would introduce excessive mass dispersion that 
the discrete execution cannot exploit. The iteration limit 
of 200 ensures convergence within numerical tolerance 
($\|\mathbf{u}^{(t+1)} - \mathbf{u}^{(t)}\|_\infty < 
10^{-5}$ in practice) for all configurations tested.

\paragraph{Software environment.}
All experiments use Python 3.10, PyTorch 2.7.0, and CUDA 12.8 
on a single NVIDIA RTX Pro 5000 (Blackwell) GPU. We use the 
lmms-eval framework~\cite{lmmseval} for both VQA and VTG 
evaluation, with the official model checkpoints for 
Qwen2.5-VL-7B~\cite{qwen2.5vl}, LLaVA-OneVision-7B~\cite{llava-ov}, and LLaVA-Video-7B~\cite{llava-video} from 
HuggingFace. All baseline methods are reproduced under the 
same evaluation protocol for fair comparison.

\paragraph{Evaluation protocol.}
VQA scores are reported as accuracy on multiple-choice 
questions, while VTG scores are reported as mean IoU (mIoU) 
between predicted and ground-truth temporal segments. 
MVBench~\cite{mvbench} consists of short videos covering 
diverse temporal reasoning skills, Video-MME~\cite{videomme} 
spans short, medium, and long videos for comprehensive 
multimodal evaluation, and MLVU-dev~\cite{mlvu} and 
LongVideoBench~\cite{longvideobench} both target long-form 
video understanding. For VTG, Charades-STA~\cite{charades} 
contains short videos captured from fixed indoor cameras, 
and ActivityNet-Captions~\cite{activitynet} (using the 
TimeLens~\cite{timelens} split) covers longer activity 
videos, though shorter than typical long-video benchmarks.

\section{Spatial Compression Analysis}
\label{appendix:spatial}

\subsection{Comparison with Alternative Spatial Methods}
\label{appendix:spatial_ablation}

We compare five spatial pruning methods within the OTT-Vid 
framework while keeping the OT-based temporal compression 
fixed. Each method consists of (i) a token selection 
algorithm and (ii) a mass function that maps each selected 
token to its preservation priority for the temporal stage. 
For fair comparison, the mass function of each method mirrors 
its selection criterion: tokens scoring high under the 
selection objective receive low mass (preserved) and 
low-scoring tokens receive high mass (compressed), with all 
methods using the same negative softmax temperature 
$\tau_m{=}0.3$.

\paragraph{Top-K.}
Selection ranks tokens by their saliency $w_j$ and retains 
the top $K$. Mass is assigned directly from saliency: 
$m_j \propto \exp(-w_j / \tau_m)$. Diversity or coverage are not 
considered, so selected tokens may concentrate within a few 
semantic clusters. 

\paragraph{DivPrune~\cite{divprune}.}
Selection uses max-min farthest-point sampling on the cosine 
distance matrix without considering saliency: each subsequent 
token maximizes its minimum cosine distance to the already 
selected set. Mass is derived from the same isolation 
quantity that drives selection, $\text{iso}_j = \min_{k \in 
S \setminus \{j\}} (1 - \cos(j, k))$, with 
$m_j \propto \exp(-\text{iso}_j / \tau_m)$.

\paragraph{ADTS~\cite{flashvid}.}
The spatial pruning component of FlashVID combines diversity 
and saliency through saliency-scaled distances: each 
candidate's distance to the selected set is multiplied by 
its saliency before max-min selection. Mass uses the same 
saliency-scaled isolation, $\text{iso}_j = w_j \cdot 
\min_{k \in S \setminus \{j\}} (1 - \cos(j, k))$.

\paragraph{SCOPE~\cite{scope}.}
Selection performs greedy coverage maximization with a coverage 
gain weighted by the candidate's own saliency: 
$\text{gain}(j) = w_j \cdot \sum_i \max(0, \cos(i,j) - 
\text{cur\_max}_i)$. Mass uses a leave-one-out form of the 
same expression, measuring how much coverage is lost when 
$j$ is removed from the selected set, scaled by $w_j$.

\paragraph{OTT-Vid (Ours).}
Selection greedily picks tokens that maximize a coverage 
gain weighted by the saliency of the covered tokens 
rather than the candidate: $\text{gain}(j) = \sum_i w_i 
\cdot \max(0, \cos(i,j) - \text{cur\_max}_i)$. While SCOPE~\cite{scope} incorporates saliency explicitly via the candidate weight $w_j$, our formulation incorporates it 
implicitly through the covered tokens. As a result, 
the iterative selection naturally prioritizes salient 
regions in the early steps where their weighted gain is 
largest, and shifts toward under-covered regions in later 
steps as the salient regions saturate. Mass follows the 
same leave-one-out form, measuring the loss in coverage of 
salient tokens when $j$ is removed.

\begin{table*}[t]
\centering
\caption{Comparison of spatial pruning methods within the 
OTT-Vid framework on Qwen2.5-VL-7B at 10\% retention. 
The temporal compression stage is fixed; only the spatial 
pruning method is varied. VQA scores are reported as 
accuracy and VTG scores as mIoU. Avg.\ (\%) denotes the mean 
performance retention relative to the uncompressed baseline. 
Best results are in \textbf{bold}.}
\label{tab:spatial_ablation}
\resizebox{\textwidth}{!}{
\setlength{\tabcolsep}{3pt}
\renewcommand{\arraystretch}{1}
\begin{tabular}{l | c | cccc | c || cc | c}
\toprule
& & \multicolumn{5}{c||}{\textbf{Video Question Answering}} 
& \multicolumn{3}{c}{\textbf{Video Temporal Grounding}} \\
\cmidrule(lr){3-7} \cmidrule(lr){8-10}
\multirow{-2}{*}{\makecell[c]{\textbf{Spatial Pruning} 
\\ \textbf{Method}}}
& \multirow{-2}{*}{\textbf{Ret.}}
& MVBench & VideoMME & LVB & MLVU & Avg. (\%)
& Charades & ANet & Avg. (\%) \\
\midrule
Saliency & 10\% 
& 63.9 & 58.5 & 56.0 & 60.0 & 95.1
& \textbf{32.4} & 16.9 & 71.5 \\
DivPrune~\cite{divprune} {\scriptsize CVPR'25} & 10\% 
& 61.9 & 56.7 & 57.2 & 59.2 & 93.7
& 30.1 & 16.6 & 68.0 \\
SCOPE~\cite{scope} {\scriptsize NeurIPS'25} & 10\% 
& 63.4 & 58.7 & \textbf{56.8} & 60.3 & 95.3
& 31.4 & 17.5 & 71.2 \\
ADTS~\cite{flashvid} {\scriptsize ICLR'26} & 10\% 
& \textbf{64.1} & \textbf{58.9} & 55.7 & 60.5 & 95.4
& 32.1 & 17.8 & 72.6 \\
\rowcolor{bestrow}
\textbf{OTT-Vid} & 10\% 
& 64.0 & 58.7 & 56.7 & \textbf{60.7} & \textbf{95.8}
& 32.3 & \textbf{18.4} & \textbf{73.9} \\
\bottomrule
\end{tabular}
}
\end{table*}

\begin{figure*}[t]
\centering
\includegraphics[width=\textwidth]{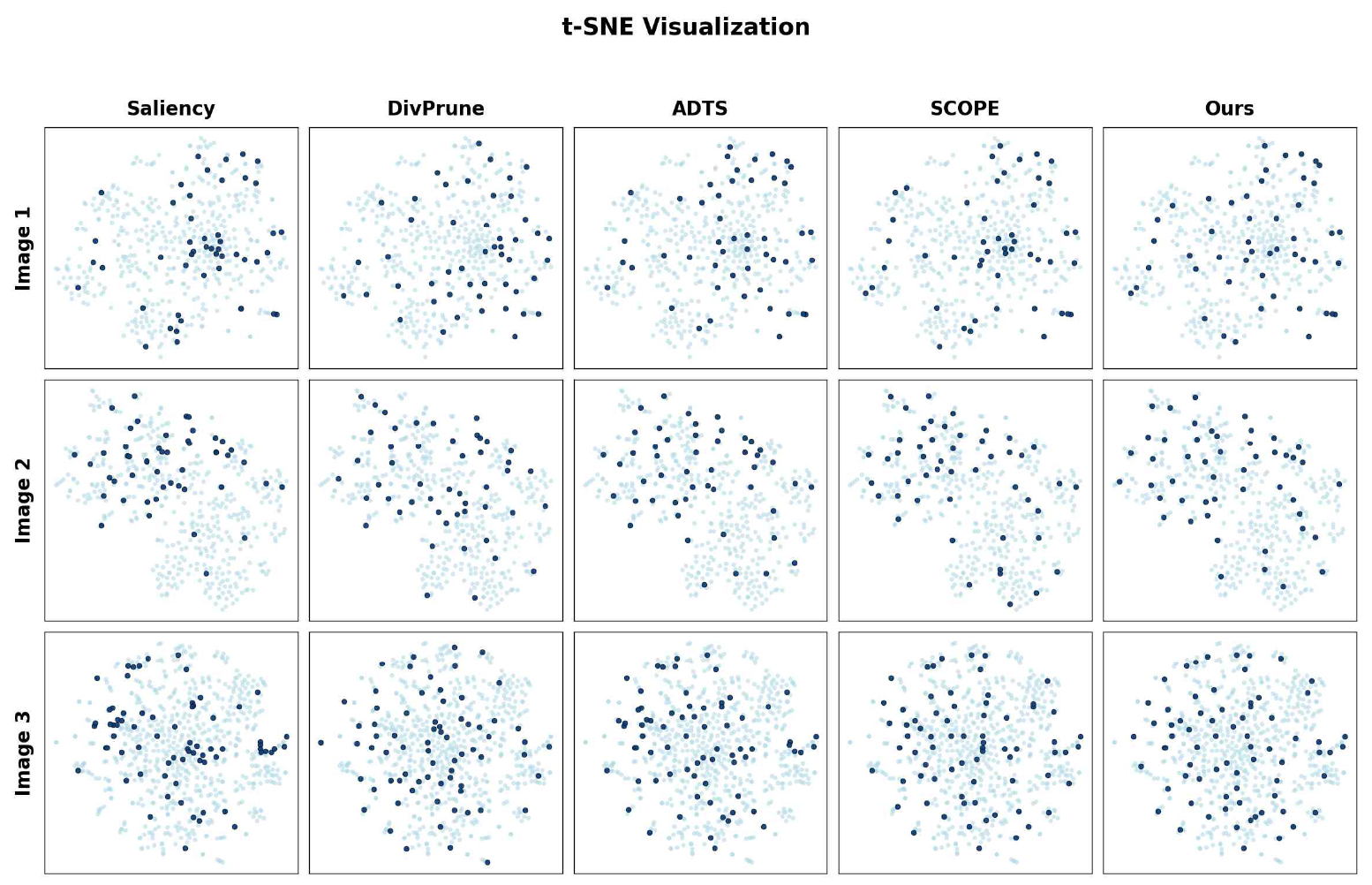}
\caption{t-SNE visualization of selected tokens (dark blue) 
versus all tokens (light blue) for the five spatial pruning 
methods on three images at $r{=}10\%$ retention on 
Qwen2.5-VL. Top-K concentrates within a few feature-space 
clusters, DivPrune spreads uniformly without prioritization, 
and our saliency-and-coverage formulation produces a 
selection that spans the full distribution while maintaining 
density on salient regions.}
\label{fig:spatial_tsne}
\end{figure*}

\paragraph{Comparison Result.}Table~\ref{tab:spatial_ablation} reports the quantitative 
results at $r{=}10\%$ retention on Qwen2.5-VL-7B, while 
Figure~\ref{fig:spatial_tsne} qualitatively illustrates 
which tokens each method retains via t-SNE. 
At the quantitative level, our saliency-and-coverage 
formulation achieves the strongest average performance on 
both task families (95.8\% VQA and 73.9\% VTG). 
On VQA, the differences across methods are modest, with 
SCOPE~\cite{scope}, FlashVID~\cite{flashvid} (ADTS), and our 
method all falling within 0.6pp of each other. On VTG, the 
gaps widen, with our method exceeding ADTS by 1.3pp and 
SCOPE by 2.7pp, suggesting that spatial pruning choice 
matters more for tasks that require fine-grained temporal 
evidence. Across both task families, the OT-based temporal 
compression remains effective regardless of the specific 
spatial choice. 

This modularity has a notable practical 
implication: when ADTS is used as the spatial stage within 
our framework, it achieves 95.4\% VQA, substantially exceeding 
the 93.9\% reported in Table~\ref{tab:main} for FlashVID~\cite{flashvid} 
as a complete pipeline at the same retention ratio. The 
temporal compression mechanism therefore contributes 
meaningfully on top of reasonable spatial pruning 
choice. The qualitative t-SNE visualization further reveals 
the design intent of each method: Saliency concentrates the 
selection within a few feature-space clusters, while 
DivPrune~\cite{divprune} spreads it uniformly without prioritization. The remaining methods (SCOPE~\cite{scope}, ADTS~\cite{flashvid}, ours) jointly consider saliency together with either diversity or coverage, producing intermediate selection patterns that combine 
clustering on salient regions with broader coverage of the 
token distribution.

\section{Adaptive Behavior Analysis}
\label{appendix:adaptive}

\subsection{Budget Feasibility and Overflow Analysis}
\label{appendix:budget_overflow}

The pairwise budget allocation distributes the global budget 
$B_{\text{tot}}$ across $T{-}1$ adjacent frame pairs via a 
softmax over transport difficulties, with no inherent upper 
bound on the per-pair budget $B_t$. However, each pair 
$(t, t{+}1)$ supports at most $K$ compression operations under 
the constraint that each source token in $S_{t+1}$ is selected 
at most once, giving $B_t \le K$. We refer to the condition 
$B_t > K$ as \emph{budget overflow}, in which the allocation 
exceeds the per-pair feasibility cap and the excess operations 
cannot be executed. In this subsection, we describe the 
mechanism handling overflow, derive its theoretical condition, 
and empirically verify its frequency under our default 
configuration.

\paragraph{Theoretical overflow condition.}
The uniform per-pair budget is 
$B_{\text{uniform}} = B_{\text{tot}} / (T-1) = K T (1-r_t) / (T-1)$. 
Defining the budget ratio $\rho_t := B_t / B_{\text{uniform}}$, 
the overflow condition $B_t > K$ becomes
\begin{equation}
\rho_t > \rho^* := \frac{T-1}{T(1-r_t)},
\label{eq:overflow_threshold}
\end{equation}
which converges to $1/(1-r_t)$ as $T \to \infty$. For our 
default configuration on Qwen2.5-VL~\cite{qwen2.5vl} ($r{=}0.10$, $\gamma{=}0.3$, $T{=}32$), 
$r_t = r^\gamma \approx 0.501$ yields $\rho^* \approx 1.94$, 
meaning a pair must receive nearly twice the uniform budget 
before exceeding the feasibility cap.

\subsection{Per-Benchmark Budget Distribution}
\label{appendix:budget_per_benchmark}

\begin{figure*}[t]
\centering
\includegraphics[width=\textwidth]{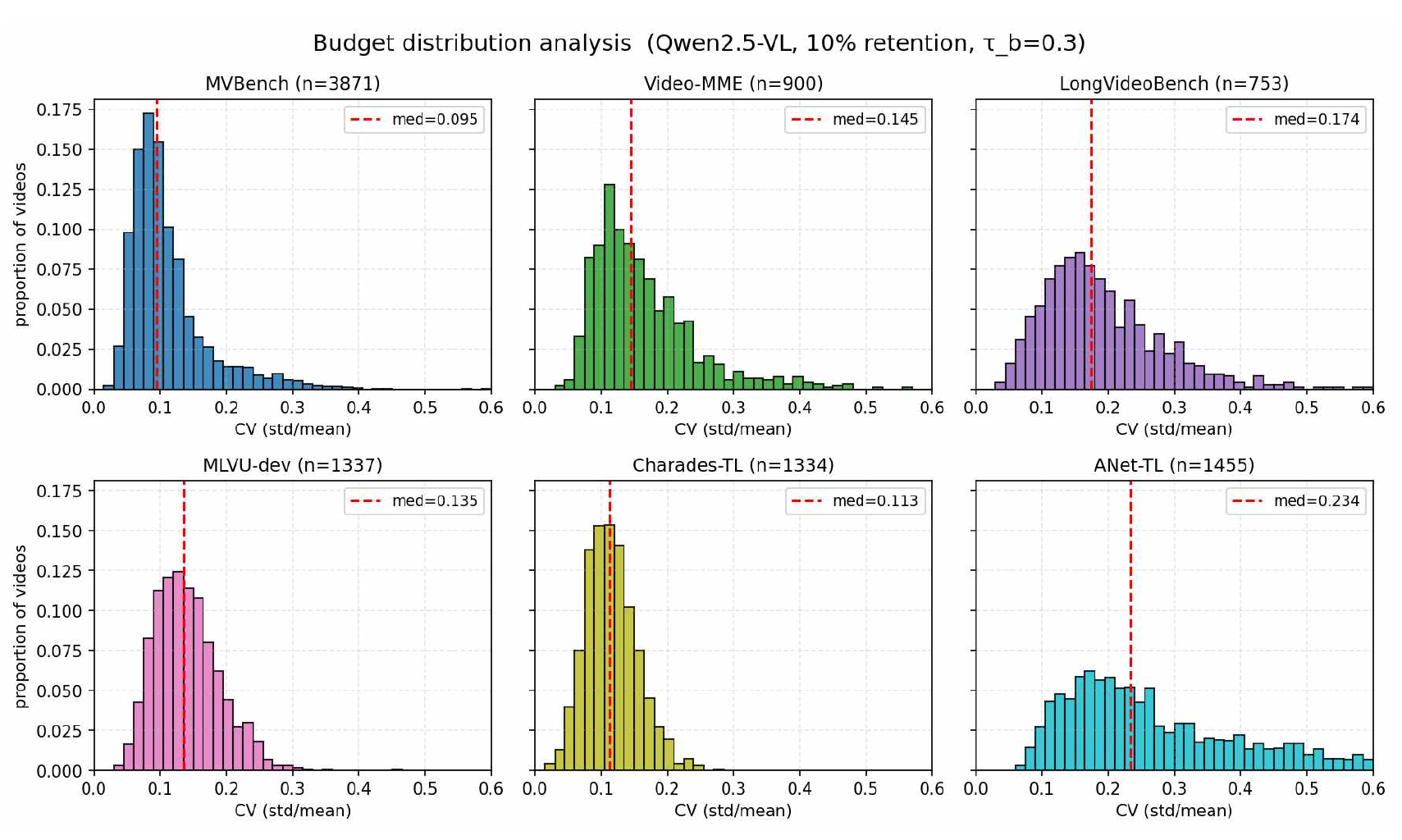}
\label{fig:benchmark_dist}
\caption{Per-video coefficient of variation 
$\text{CV} = \sigma(B_t)/\mu(B_t)$ of pairwise budgets across 
six benchmarks on Qwen2.5-VL at $r{=}0.10$ and $\tau_b{=}0.3$. 
Red dashed lines mark medians.}
\label{fig:benchmark_cv}
\end{figure*}

\paragraph{Per-benchmark budget variability.}
To characterize how non-uniformly OTT-Vid distributes the 
budget within a single video, we compute the coefficient of 
variation $\text{CV} = \sigma(B_t) / \mu(B_t)$, where $\sigma$ 
and $\mu$ denote the standard deviation and mean of the 
pairwise budgets $\{B_t\}_{t=1}^{T-1}$. A CV of 0 corresponds 
to a uniform allocation, while larger CV indicates that the 
adaptive mechanism concentrates budget on a smaller subset of 
frame pairs. Figure~\ref{fig:benchmark_cv} reports the 
distribution of per-video CV across the six benchmarks on 
Qwen2.5-VL at $r{=}0.10$ and $\tau_b{=}0.3$. The median CV 
ranges from 0.095 (MVBench) to 0.234 (ANet-TL), reflecting 
substantial differences in the temporal compressibility of 
each benchmark. MVBench and Charades-TL exhibit the lowest 
median CV (0.095 and 0.113), with most videos producing 
nearly uniform budget allocations. MLVU-dev (0.135) and 
Video-MME (0.145) form an intermediate group, while 
LongVideoBench (0.174) and ANet-TL (0.234) show substantially 
higher variability, with a long tail of videos exceeding CV 
0.4.

\subsection{Per-LLM Budget Distribution}
\label{appendix:budget_per_llm}

\begin{figure*}[t]
\centering
\includegraphics[width=\textwidth]{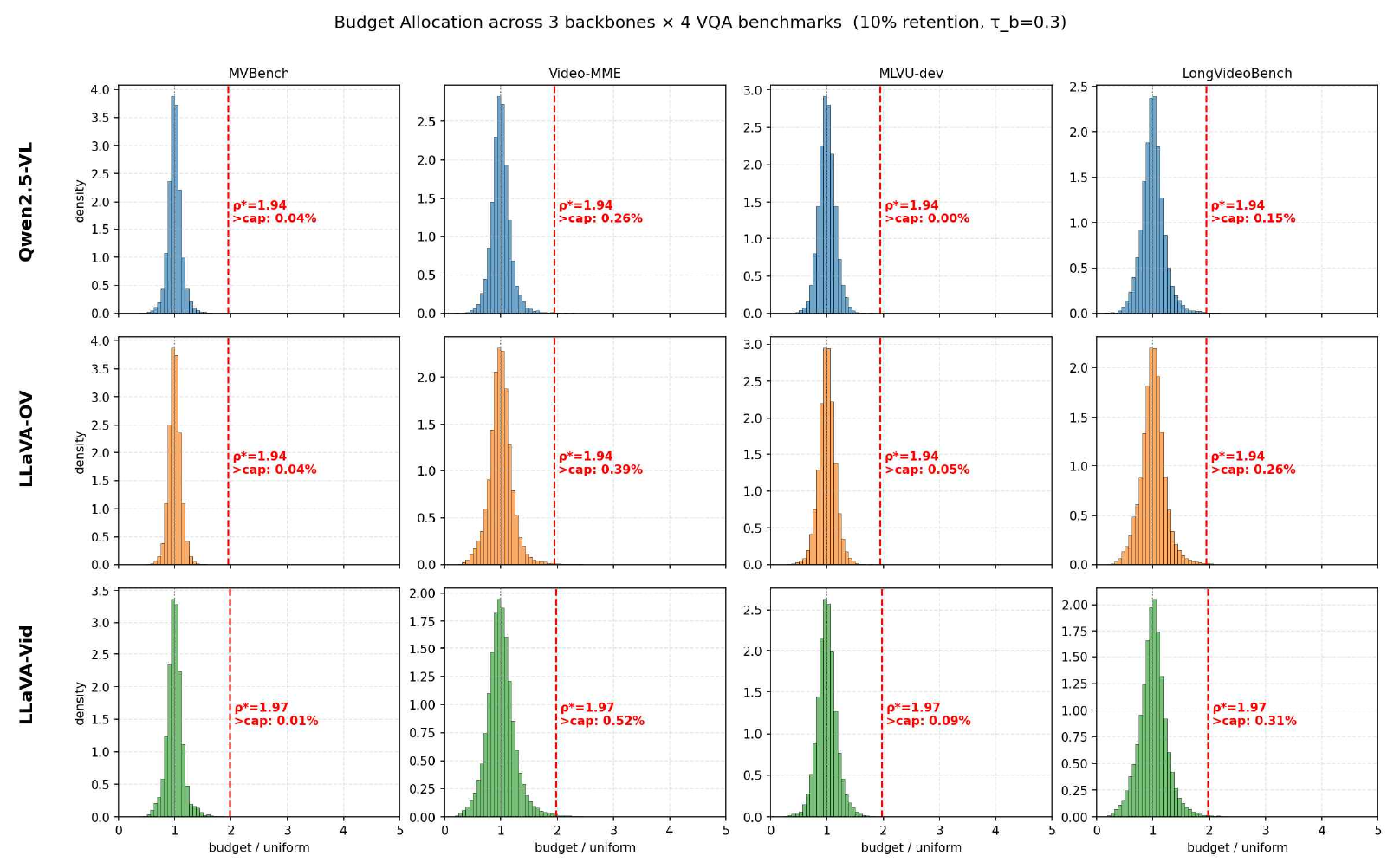}
\caption{Pooled distribution of pair-level budget ratios 
$\rho_t = B_t / B_{\text{uniform}}$ across three Video-LLM 
backbones (rows) and four VQA benchmarks (columns) at 
$r{=}0.10$ and $\tau_b{=}0.3$. Red dashed lines mark the 
overflow threshold $\rho^*$, with annotated overflow rates 
(\texttt{>cap}). The slight variation in $\rho^*$ between 
backbones reflects different frame counts $T$ (32 for 
Qwen2.5-VL and LLaVA-OV, 64 for LLaVA-Video). All twelve 
combinations show overflow rates below 1\%.}
\label{fig:llm_budget}
\end{figure*}

\paragraph{Cross-backbone budget distribution and overflow.}
While CV captures within-video variability, we further examine 
the pair-level distribution of budgets across all videos and 
backbones. Specifically, we measure the ratio 
$\rho_t = B_t / B_{\text{uniform}}$ for each frame pair, where 
$\rho_t {=} 1$ indicates a budget equal to the uniform 
allocation, $\rho_t {<} 1$ indicates a pair receiving less than 
its uniform share, and $\rho_t {>} 1$ indicates a pair 
receiving more. Figure~\ref{fig:llm_budget} shows the pooled 
distribution of $\rho_t$ across three Video-LLM backbones 
(Qwen2.5-VL, LLaVA-OV, LLaVA-Video) and four VQA benchmarks at 
$r{=}0.10$ and $\tau_b{=}0.3$. The red dashed line marks the 
overflow threshold $\rho^*$ derived in 
Eq.~\eqref{eq:overflow_threshold}: $\rho^* {=} 1.94$ for 
Qwen2.5-VL and LLaVA-OV ($T{=}32$), and $\rho^* {=} 1.97$ for 
LLaVA-Video ($T{=}64$). Across all twelve backbone-benchmark 
combinations, the empirical overflow rate remains below 1\%, 
with a maximum of 0.52\% (LLaVA-Video on Video-MME). The vast 
majority of pairs fall within $[0.5, 1.5]$, well inside the 
feasibility region. This confirms that the softmax allocation 
naturally produces feasible budgets across diverse video 
distributions and backbone configurations, and the redistribution 
mechanism described above is rarely activated in practice.

\paragraph{Handling overflow.}
During budget allocation process, if the initial softmax produces $B_t > K$ for some pair, the excess is redistributed through an iterative procedure that recomputes the softmax over the active set of pairs below their cap. The allocated amount is applied up to 
the headroom $K - B_t$ of each pair, with any excess carried 
to the next iteration. Pairs reaching $K$ are frozen and 
excluded thereafter. This preserves the softmax-based 
prioritization within the feasible region and guarantees 
$\sum_t B_t = B_{\text{tot}}$ exactly. As 
Figure~\ref{fig:llm_budget} shows, overflow is rare in 
practice, and most allocations terminate after the initial 
softmax step on the full set.

\subsection{Budget Allocation Across Time}
\label{appendix:budget_visualization}

\begin{figure*}[t]
\centering
\includegraphics[width=\textwidth]{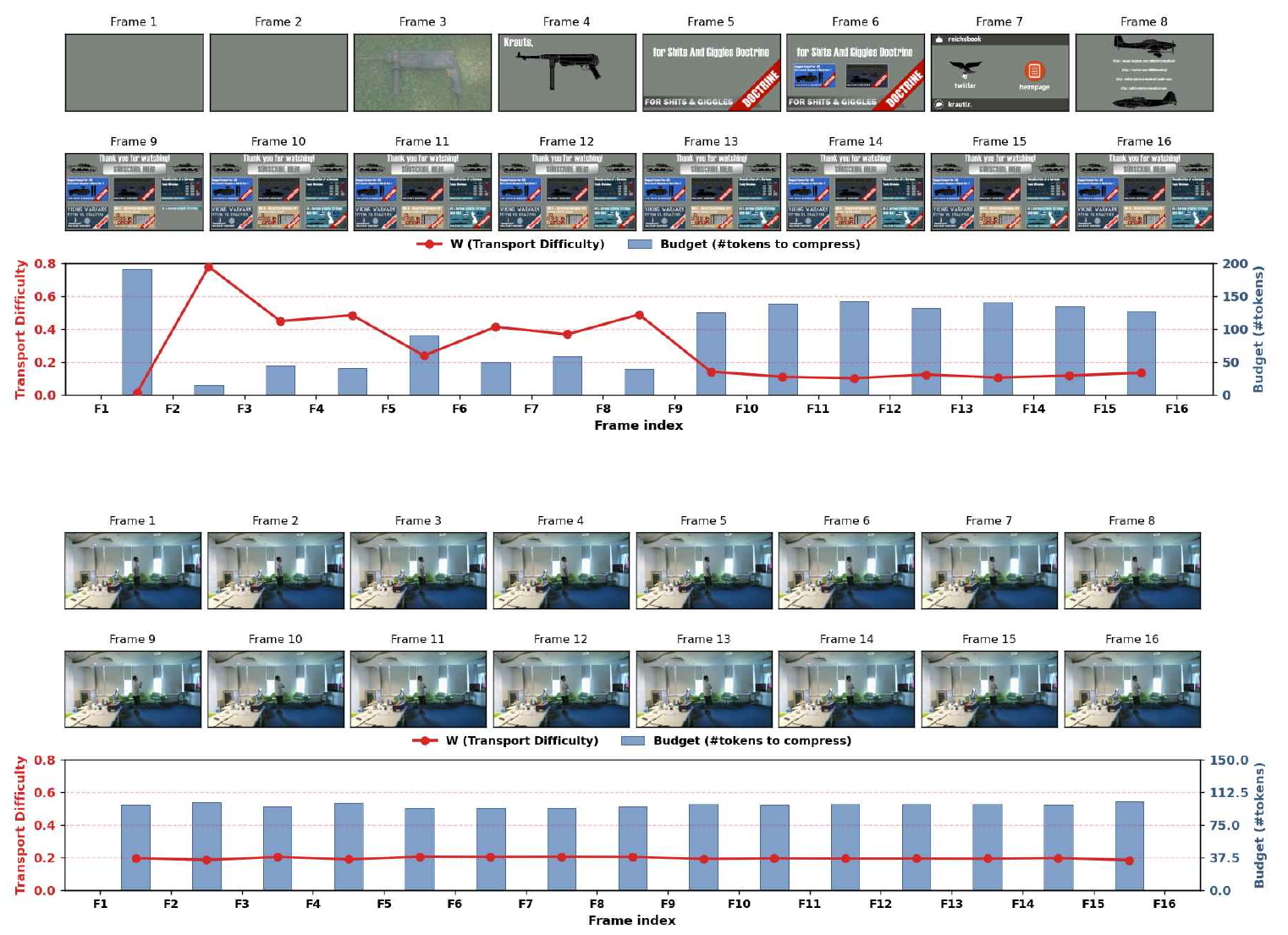}
\caption{Per-pair transport difficulty $W_t$ (red) and 
allocated budget $B_t$ (blue) on 10\% retention at 
Qwen2.5-VL with $T{=}16$. \textbf{Top:} dynamic video. 
\textbf{Bottom:} static video.}
\label{fig:budget_visualization}
\end{figure*}

Figure~\ref{fig:budget_visualization} visualizes the 
per-pair transport difficulty $W_t$ (red line) and the 
allocated budget $B_t$ (blue bars) on two videos with 
contrasting temporal characteristics. In the dynamic video 
(top), where frame content changes substantially across 
pairs, $W_t$ varies accordingly and the budget is allocated 
non-uniformly, concentrating on pairs with smaller content 
changes. In the static video (bottom), where consecutive 
frames depict nearly identical scenes, $W_t$ remains 
uniformly low and the budget distribution approaches the 
uniform allocation. The adaptive mechanism thus produces 
non-uniform budgets when warranted by the video content and 
near-uniform budgets otherwise.

\section{Additional Ablation Study}
\label{appendix:ablation}

\begin{table*}[t]
\centering
\caption{Per-benchmark results of the core component ablation 
on Qwen2.5-VL-7B at 10\% retention. Mass denotes 
importance-aware mass (\checkmark) vs.\ uniform. Budget 
denotes adaptive allocation (\checkmark) vs.\ uniform. 
VQA scores are reported as accuracy and VTG scores as mIoU. 
Avg.\ (\%) denotes the mean performance retention relative 
to the uncompressed baseline.}
\label{tab:ablation_per_benchmark}
\resizebox{\textwidth}{!}{
\setlength{\tabcolsep}{6pt}
\renewcommand{\arraystretch}{1}
\begin{tabular}{c | cc | cccc | c || cc | c}
\toprule
& \multicolumn{2}{c|}{\textbf{Comp.}} 
& \multicolumn{5}{c||}{\textbf{Video Question Answering}} 
& \multicolumn{3}{c}{\textbf{Video Temporal Grounding}} \\
\cmidrule(lr){2-3} \cmidrule(lr){4-8} \cmidrule(lr){9-11}
\multirow{-2}{*}{\textbf{Row}}
& Mass & Budget
& MVBench & VideoMME & LVB & MLVU & Avg. (\%)
& Charades & ANet & Avg. (\%) \\
\midrule
(a) & & 
& 63.7 & 57.5 & 55.6 & 59.7 & 94.3
& 30.2 & 13.5 & 63.1 \\
(b) & & \checkmark
& 63.2 & 57.7 & 56.5 & 59.7 & 94.6 
& 30.5 & 14.0 & 64.3 \\
(c) & \checkmark &
& 63.9 & 58.3 & 56.0 & 60.4 & 95.2
& \textbf{32.0} & 17.6 & 72.2 \\
\rowcolor{bestrow}
(d) & \checkmark & \checkmark
& \textbf{64.0} & \textbf{58.7} & \textbf{56.7} & \textbf{60.7} & \textbf{95.8}
& \textbf{32.3} & \textbf{18.4} & \textbf{73.9} \\
\bottomrule
\end{tabular}
}
\end{table*}

\paragraph{Per-benchmark results of core component ablation.}
Table~\ref{tab:ablation_per_benchmark} provides the 
per-benchmark breakdown of the core component ablation 
summarized in Table~\ref{tab:ablation}. The results show 
that mass and budget contributions vary across benchmarks, 
consistent with the analyses of $\tau_m$ and $\tau_b$ in 
the following subsections.

\subsection{Budget Temperature \texorpdfstring{$\tau_b$}{tau\_b}}
\label{appendix:tau_b}

\begin{table*}[t]
\centering
\caption{Effect of the budget temperature $\tau_b$ on Qwen2.5-VL-7B 
at 10\% retention. VQA scores are reported as accuracy and 
VTG scores as mIoU. Avg.\ (\%) denotes the mean performance 
retention relative to the uncompressed baseline.}
\label{tab:tau_b}
\resizebox{\textwidth}{!}{
\setlength{\tabcolsep}{6pt}
\renewcommand{\arraystretch}{1}
\begin{tabular}{c | c | cccc | c || cc | c}
\toprule
& & \multicolumn{5}{c||}{\textbf{Video Question Answering}} 
& \multicolumn{3}{c}{\textbf{Video Temporal Grounding}} \\
\cmidrule(lr){3-7} \cmidrule(lr){8-10}
\multirow{-2}{*}{\textbf{$\tau_b$}}
& \multirow{-2}{*}{\textbf{Ret.}}
& MVBench & VideoMME & LVB & MLVU & Avg. (\%)
& Charades & ANet & Avg. (\%) \\
\midrule
0.1 & 10\% 
& \textbf{64.2} & 57.8 & 56.2 & \textbf{60.9} & 95.3
& 32.1 & 18.0 & 73.0 \\
\rowcolor{bestrow}
\textbf{0.3} & 10\% 
& 64.0 & \textbf{58.7} & \textbf{56.7} & 60.7 & \textbf{95.8}
& \textbf{32.3} & \textbf{18.4} & \textbf{73.9} \\
1.0 & 10\% 
& 64.1 & 58.2 & 56.2 & 60.5 & 95.3 
& \textbf{32.3} & 17.9 & 73.1 \\
$\infty$ & 10\% 
& 63.9 & 58.3 & 56.0 & 60.4 & 95.2
& 32.0 & 17.6 & 72.2 \\
\bottomrule
\end{tabular}
}
\end{table*}

The budget temperature $\tau_b$ controls the sharpness of the 
softmax distribution over transport difficulties $\{W_t\}$ in 
Eq.~\eqref{eq:budget_allocation}. As $\tau_b \to 0$, the 
allocation concentrates on the pair with the smallest $W_t$, 
amplifying the adaptive behavior but pushing the budget ratio 
$\rho_t$ closer to the overflow threshold $\rho^*$. As 
$\tau_b \to \infty$, the allocation converges to the uniform 
baseline $B_t = B_{\text{uniform}}$, eliminating the 
data-dependent prioritization. Intermediate values balance 
these two regimes by modulating how strongly the relative 
differences in $\{W_t\}$ influence the budget distribution.

Table~\ref{tab:tau_b} reports the effect of $\tau_b$ on 
average performance at $r{=}0.10$ on Qwen2.5-VL-7B. As 
$\tau_b \to \infty$, the allocation converges to the uniform 
baseline and performance degrades on both task families, 
confirming that adaptive allocation contributes meaningfully 
beyond uniform budgeting. Conversely, smaller $\tau_b$ 
sharpens the allocation but increases overflow frequency, 
triggering the iterative redistribution more often. The 
default $\tau_b{=}0.3$ operates in the intermediate region 
where overflow remains below 1\% across all benchmarks 
while preserving meaningful non-uniformity in the allocation.

The sensitivity to $\tau_b$ also varies across benchmarks 
in a way consistent with their compressibility profiles in 
Figure~\ref{fig:benchmark_cv}. Benchmarks with low median 
CV, such as MVBench and Charades, show minimal performance 
variation across $\tau_b$, as their nearly uniform budget 
distributions leave little room for $\tau_b$ to amplify. 
In contrast, benchmarks with higher CV, such as ANet, are 
more sensitive to $\tau_b$ and peak around the default 
$\tau_b{=}0.3$. This indicates that $\tau_b$ plays a more substantive role 
on videos with dynamic temporal context, where compressibility 
varies across frame pairs, while having limited effect on 
videos with more uniform temporal content.

\subsection{Mass Temperature \texorpdfstring{$\tau_m$}{tau\_m}}
\label{appendix:tau_m}

The mass temperature $\tau_m$ controls the sharpness of the 
negative softmax in Eq.~\eqref{eq:mass_assignment}, which 
converts per-token contribution scores $\tilde{u}_{t,k}$ into 
the transport mass vector $m_t$. As $\tau_m \to 0$, the mass 
distribution concentrates on tokens with the smallest 
contribution scores, sharply prioritizing semantically 
representative tokens for preservation while making the 
transport plan strongly biased against compressing them. As 
$\tau_m \to \infty$, the mass distribution converges to a 
uniform vector $m_{t,k} = 1/K$, in which case the OT marginal 
constraint no longer encodes preservation priority and the 
transport reduces to similarity-driven matching. Intermediate 
values modulate how strongly intra-frame token importance 
influences the transport plan.

\begin{table*}[t]
\centering
\caption{Effect of the mass temperature $\tau_m$ on Qwen2.5-VL-7B 
at 10\% retention. VQA scores are reported as accuracy and 
VTG scores as mIoU. Avg.\ (\%) denotes the mean performance 
retention relative to the uncompressed baseline.}
\label{tab:tau_m}
\resizebox{\textwidth}{!}{
\setlength{\tabcolsep}{6pt}
\renewcommand{\arraystretch}{1}
\begin{tabular}{c | c | cccc | c || cc | c}
\toprule
& & \multicolumn{5}{c||}{\textbf{Video Question Answering}} 
& \multicolumn{3}{c}{\textbf{Video Temporal Grounding}} \\
\cmidrule(lr){3-7} \cmidrule(lr){8-10}
\multirow{-2}{*}{\textbf{$\tau_m$}}
& \multirow{-2}{*}{\textbf{Ret.}}
& MVBench & VideoMME & LVB & MLVU & Avg. (\%)
& Charades & ANet & Avg. (\%) \\
\midrule
0.1 & 10\% 
& 63.6 & 58.1 & \textbf{56.7} & 60.3 & 95.2
& 31.6 & 17.4 & 71.3 \\
\rowcolor{bestrow}
\textbf{0.3} & 10\% 
& 64.0 & \textbf{58.7} & \textbf{56.7} & \textbf{60.7} & \textbf{95.8}
& \textbf{32.3} & \textbf{18.4} & \textbf{73.9} \\
1.0 & 10\% 
& \textbf{64.0} & 57.6 & 56.5 & 60.0 & 95.0 
& 31.8 & 16.2 & 69.6 \\
$\infty$ & 10\% 
& 63.2 & 57.7 & 56.5 & 59.7 & 94.6
& 30.5 & 14.0 & 64.3 \\
\bottomrule
\end{tabular}
}
\end{table*}

Table~\ref{tab:tau_m} reports the effect of $\tau_m$ on 
average performance at $r{=}0.10$ on Qwen2.5-VL-7B. The 
uniform-mass case ($\tau_m \to \infty$) shows a sharp drop 
on both task families and especially on VTG, confirming 
that without preservation priority, important tokens fail 
to persist across frames. At the other end, $\tau_m{=}0.1$ 
also degrades performance: the sharply peaked mass 
distribution biases the transport plan toward mass over 
cost, restricting its flexibility to find low-cost 
cross-frame correspondences. The default $\tau_m{=}0.3$ 
balances these two regimes, preserving important tokens 
through non-uniform mass while leaving sufficient room 
for cost-driven matching.

\begin{table*}[t]
\centering
\caption{Effect of the cost threshold $\tau_c$ on Qwen2.5-VL-7B 
at 10\% retention. VQA scores are reported as accuracy and 
VTG scores as mIoU. Avg.\ (\%) denotes the mean performance 
retention relative to the uncompressed baseline.}
\label{tab:tau_c}
\resizebox{\textwidth}{!}{
\setlength{\tabcolsep}{6pt}
\renewcommand{\arraystretch}{1}
\begin{tabular}{c | c | cccc | c || cc | c}
\toprule
& & \multicolumn{5}{c||}{\textbf{Video Question Answering}} 
& \multicolumn{3}{c}{\textbf{Video Temporal Grounding}} \\
\cmidrule(lr){3-7} \cmidrule(lr){8-10}
\multirow{-2}{*}{\textbf{$\tau_c$}}
& \multirow{-2}{*}{\textbf{Ret.}}
& MVBench & VideoMME & LVB & MLVU & Avg. (\%)
& Charades & ANet & Avg. (\%) \\
\midrule
0.0 & 10\% 
& 63.8 & 57.9 & 56.9 & 60.2 & 95.3
& 32.1 & 18.0 & 73.0 \\
\rowcolor{bestrow}
\textbf{0.3} & 10\% 
& \textbf{64.0} & \textbf{58.7} & 56.7 & \textbf{60.7} & \textbf{95.8}
& \textbf{32.3} & \textbf{18.4} & \textbf{73.9} \\
0.7 & 10\% 
& 63.4 & \textbf{58.7} & \textbf{57.6} & 60.6 & \textbf{95.8} 
& 32.2 & 18.1 & 73.3 \\
1.0 & 10\% 
& 63.8 & 58.1 & 57.1 & 60.1 & 95.4
& 32.1 & 17.8 & 72.6 \\
\bottomrule
\end{tabular}
}
\end{table*}

\subsection{Cost Threshold $\tau_c$}
\label{appendix:tau_c}

The cost threshold $\tau_c$ determines whether each selected 
match in the transport plan is executed as a merge (uniform 
averaging) or as a prune (deletion of the source token), 
with matches satisfying $C_{t,ij} < \tau_c$ being merged and 
the remainder pruned. We introduced this dual-mode 
formulation following the well-established principle that 
merging is reliable when matched representations are close 
enough that uniform averaging preserves their information, 
while pruning is preferable when the matched representations 
differ substantially~\cite{tofu, pact, pumer}. Recent video token compression methods such 
as DyCoke~\cite{dycoke} and 
PruneVid~\cite{prunevid} adopt similar dual-mode 
strategies, and we incorporate $\tau_c$ as a principled 
safeguard against unreliable cross-frame correspondences.

Table~\ref{tab:tau_c} reports the effect of $\tau_c$ at 
$r{=}10\%$ on Qwen2.5-VL-7B. Performance is robust to 
$\tau_c$ across a wide range, but degrades slightly at 
both extremes. At $\tau_c{=}0$ (all-pruning), the dual-mode 
formulation collapses to pure pruning, discarding 
information from reliable low-cost matches that uniform 
averaging could have preserved. At $\tau_c{=}1.0$ 
(all-merging), high-cost matches are merged despite 
representing semantically unreliable correspondences, 
contaminating the target representation through averaging. 
Within the intermediate range $\tau_c \in [0.3, 0.7]$, 
performance remains stable because high-cost matches 
typically involve source tokens with relatively low 
importance, so the choice between merging and pruning 
yields similar downstream effects. We adopt $\tau_c{=}0.3$ 
as default, which also offers compression overhead 
efficiency since pruning removes tokens directly rather 
than computing weighted averages.

\subsection{Mixing Weight $\alpha_t$}
\label{appendix:alpha}

\begin{table*}[t]
\centering
\caption{Effect of the mixing weight $\alpha_t$ formulation 
on Qwen2.5-VL-7B at 10\% retention. Fixed values 
($\alpha{=}1$, $\alpha{=}0.5$) correspond to the upper 
and lower bounds of our formulation, while the dynamic 
variants compute $\bar{s}_t$ at different spatial scales. 
VQA scores are reported as accuracy and VTG scores as mIoU. 
Avg.\ (\%) denotes the mean performance retention relative 
to the uncompressed baseline.}
\label{tab:alpha}
\resizebox{\textwidth}{!}{
\setlength{\tabcolsep}{6pt}
\renewcommand{\arraystretch}{1}
\begin{tabular}{l | c | cccc | c || cc | c}
\toprule
& & \multicolumn{5}{c||}{\textbf{Video Question Answering}} 
& \multicolumn{3}{c}{\textbf{Video Temporal Grounding}} \\
\cmidrule(lr){3-7} \cmidrule(lr){8-10}
\multirow{-2}{*}{\textbf{Variant}}
& \multirow{-2}{*}{\textbf{Ret.}}
& MVBench & VideoMME & LVB & MLVU & Avg. (\%)
& Charades & ANet & Avg. (\%) \\
\midrule
$\alpha{=}1$       & 10\% & 63.8 & 58.1 & 55.9 & 60.5 & 95.0 & 32.1 & 18.0 & 73.0 \\
$\alpha{=}0.5$     & 10\% & 63.8 & 58.4 & 56.3 & 59.9 & 95.1 & 32.0 & 18.1 & 73.0 \\
\midrule
$\alpha{=}$global  & 10\% & \textbf{64.3} & \textbf{59.0} & 56.6 & 60.4 & \textbf{95.8} & 32.1 & 18.2 & 73.3 \\
$\alpha{=}$kernel  & 10\% & 64.2 & 58.6 & 56.3 & 60.6 & 95.6 & \textbf{32.4} & \textbf{18.5} & \textbf{74.2} \\
\rowcolor{bestrow}
\textbf{Ours} & 10\% 
& 64.0 & 58.7 & \textbf{56.7} & \textbf{60.7} & \textbf{95.8} & 32.3 & 18.4 & 73.9 \\
\bottomrule
\end{tabular}
}
\end{table*}

We compare five formulations of the mixing weight $\alpha_t$ 
in Eq.~\eqref{eq:cost_matrix}: two fixed values 
($\alpha{=}1$ and $\alpha{=}0.5$, corresponding to the upper 
and lower bounds in our formulation), and three dynamic 
variants based on $\bar{s}_t$ computed at different spatial 
scales: position-aligned (per-token similarity at identical 
grid positions, our default), kernel-based ($3{\times}3$ 
neighborhood averaging without padding), and global 
(frame-level mean similarity).

Table~\ref{tab:alpha} reports the results at $r{=}0.10$ on 
Qwen2.5-VL-7B. The dynamic variants consistently outperform 
the fixed values, confirming that adapting $\alpha_t$ to 
scene dynamics is beneficial. Among the dynamic variants, 
position-aligned, kernel, and global perform comparably on 
VQA benchmarks, but global is weaker on VTG. We attribute 
this to the difference in frame sampling: VQA's sparse 
32-frame sampling produces large frame-to-frame changes 
captured even at coarse spatial resolutions, while VTG's 
dense 2~fps sampling yields small, spatially localized 
differences between adjacent frames. Under such fine 
granularity, the global variant's frame-level averaging 
smooths out localized motion and loses the ability to 
distinguish static and dynamic regions within a pair. The 
position-aligned and kernel variants retain per-token (or 
near per-token) spatial resolution and remain sensitive to 
localized dynamics, yielding consistent performance across 
both task families. We adopt position-aligned as default 
for its simplicity, since per-token similarity already 
captures the local correspondence signal that kernel 
averaging would aggregate, making the additional 
neighborhood operation unnecessary in practice.

\label{appendix:alpha}

\section{Additional Qualitative Results}
\label{appendix:visualize}

\begin{figure}[t]
\centering
\includegraphics[width=\linewidth]{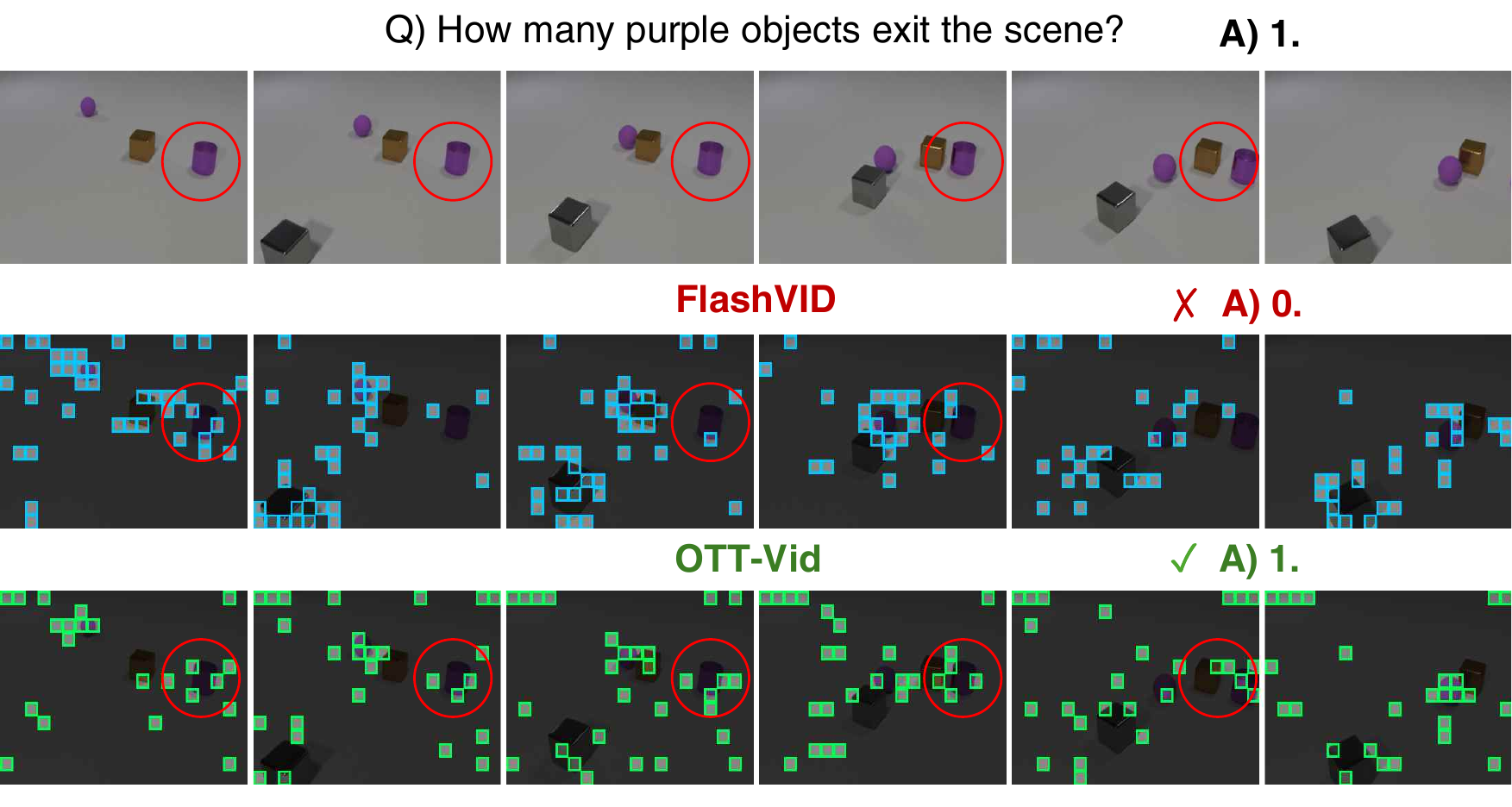}
\caption{Qualitative comparison on MVBench at 10\% retention. 
Bright patches indicate retained tokens. Red circles highlight 
the question-relevant region.}
\label{fig:qualitative_mvbench}
\end{figure}

\begin{figure}[t]
\centering
\includegraphics[width=\linewidth]{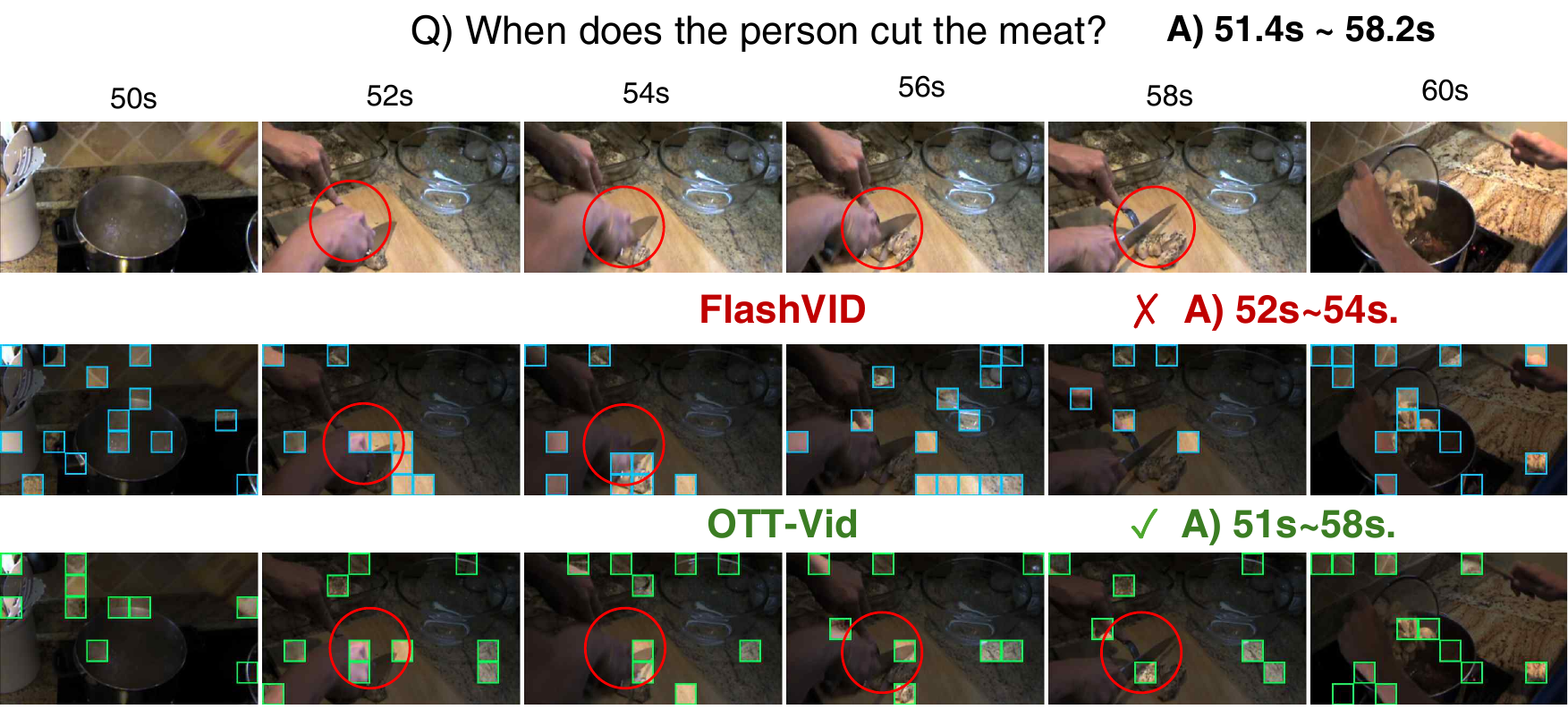}
\caption{Qualitative comparison on ActivityNet-TimeLens at 
10\% retention. Bright patches indicate retained tokens. 
Red circles highlight the question-relevant region.}
\label{fig:qualitative_anet}
\end{figure}

We provide additional qualitative comparisons with FlashVID~\cite{flashvid} 
on MVBench (Figure~\ref{fig:qualitative_mvbench}) and 
ActivityNet-TimeLens (Figure~\ref{fig:qualitative_anet}) at 
10\% retention. In both figures, bright patches indicate 
retained tokens and red circles mark the query-relevant 
regions.

Figure~\ref{fig:qualitative_mvbench} shows a counting query 
asking how many purple objects exit the scene. The purple 
cylinder on the right gradually exits as the camera moves, 
serving as the key evidence for the correct answer. FlashVID's 
similarity-driven temporal compression treats this object as 
redundant across frames and fails to retain tokens covering 
its exit, leading to an incorrect count of zero. OTT-Vid 
preserves tokens around the cylinder throughout the sequence 
under its importance-aware mass formulation, capturing the 
exit evidence and producing the correct answer.

Figure~\ref{fig:qualitative_anet} shows a temporal grounding 
query for the moment of cutting the meat (ground truth: 
51.4--58.2s). The cutting action repeats across multiple 
frames in this interval, and FlashVID's similarity-based 
compression treats the recurring motion as redundant, 
discarding tokens around the hand and knife in the middle 
frames (e.g., 56s). This loss collapses the localized 
estimate to 52--54s, missing most of the action span. 
OTT-Vid retains the cutting region across all frames in 
the interval, producing a temporal estimate (51--58s) 
that closely matches the ground truth.

\vspace{-2cm}

\clearpage

\end{document}